%% file: tang-wan-liao21.tex
\documentclass[review]{elsarticle}
\usepackage{amsthm}
\usepackage{lineno,hyperref}
\usepackage{moreverb}
\usepackage{graphicx,subfig,wrapfig}
\usepackage{algorithm}
\usepackage{algpseudocode}
\usepackage{color}
\usepackage{amsmath}
\usepackage{txfonts}
\usepackage{afterpage}
\usepackage[margin=1in]{geometry}
\usepackage{mathrsfs}
\usepackage{tikz,amsmath}
\usepackage{pstricks}
\usepackage{diagbox}
\usepackage{arydshln}
\usepackage{dsfont}
\usepackage{smartdiagram}
\usepackage[normalem]{ulem}
\usepackage{todonotes}

\usepackage{verbatim}

\usetikzlibrary{calc,trees,positioning,arrows,chains,shapes.geometric,%
    decorations.pathreplacing,decorations.pathmorphing,shapes,%
    matrix,shapes.symbols}

\tikzset{
>=stealth',
  punktchain/.style={
    rectangle, 
    rounded corners, 
    draw=black, very thick,
    text width=10em, 
    minimum height=2em, 
    text centered, 
    on chain},
  line/.style={draw, thick, <-},
  element/.style={
    tape,
    top color=white,
    bottom color=blue!50!black!60!,
    minimum width=6em,
    draw=blue!40!black!90, very thick,
    text width=10em, 
    minimum height=3.5em, 
    text centered, 
    on chain},
  every join/.style={->, thick,shorten >=1pt},
  decoration={brace},
  tuborg/.style={decorate},
  tubnode/.style={midway, right=2pt},
}

\tikzstyle{arrow} = [thick,->,>=stealth]

\newdefinition{rmk}{Remark}
\newproof{pf}{Proof}

\newcommand{\bG}{\boldsymbol{G}}
\newcommand{\bx}{\boldsymbol{x}}

\newcommand{\bq}{\boldsymbol{q}}

\newcommand{\mx}{\mb{x}}

\newcommand{\bmu}{\boldsymbol{\mu}}

\newcommand{\norm}[2]{\left\| #1 \right\|_{#2}}

\newcommand{\mb}{\mathbf}

\newcommand{\xw}[1]{{\color{black}{#1}}}
\newcommand{\revs}[1]{{\color{black}{#1}}}

\renewcommand{\mathbf}{\boldsymbol}
\newcommand{\xs}{\mathbb}

\DeclareMathOperator*{\argmin}{arg\,min}

\newcommand{\comm}[1]{}

\makeatletter
\def\ps@pprintTitle{%
   \let\@oddhead\@empty
   \let\@evenhead\@empty
   \let\@oddfoot\@empty
   \let\@evenfoot\@oddfoot
}
\makeatother

\modulolinenumbers[5]

\journal{Journal of Computational Physics}

\begin{document}

\begin{frontmatter}
		
		\title{Adaptive deep density approximation for Fokker-Planck equations}
		
		
		\author[mymainaddress,mysecondaryaddress]{Kejun Tang}
		\ead{tangkj@shanghaitech.edu.cn}
		
		\author[mythirdaddress]{Xiaoliang Wan}
		\ead{xlwan@lsu.edu}
		
		\author[mymainaddress]{Qifeng Liao\corref{mycorrespondingauthor}}
		\cortext[mycorrespondingauthor]{Corresponding author}
		\ead{liaoqf@shanghaitech.edu.cn}
		
		\address[mymainaddress]{School of Information Science and Technology, ShanghaiTech University, Shanghai 201210, China}
		\address[mysecondaryaddress]{Peng Cheng Laboratory, Shenzhen 518000, China}
		\address[mythirdaddress]{Department of Mathematics and Center for Computation 
			and Technology, 
			Louisiana State University, Baton Rouge 70803, USA}

		\begin{abstract}
In this paper we present an 
adaptive deep density approximation strategy based on KRnet (ADDA-KR) 
for solving the steady-state Fokker-Planck (F-P) equations. \xw{F-P equations are usually high-dimensional and defined on an unbounded domain,} 
which limits the application of traditional grid based numerical methods. 
With the Knothe-Rosenblatt rearrangement, our newly proposed flow-based generative model, called KRnet, 
provides a family of probability density functions to serve as effective solution candidates 
for the Fokker-Planck equations, which has a weaker dependence on dimensionality than traditional computational approaches and \xw{can efficiently estimate general high-dimensional density functions.}
To obtain effective stochastic collocation points for \xw{the approximation of the F-P equation}, we develop an adaptive sampling procedure,
where samples are generated iteratively using \xw{the approximate density function} at each iteration.
We present a general 
framework of ADDA-KR, 
validate its accuracy and demonstrate its efficiency with numerical experiments.

		\end{abstract}
		
		\begin{keyword}
			density estimation; flow-based generative models; Fokker-Planck equations; deep learning. 
		\end{keyword}
		
	\end{frontmatter}

\section{Introduction}\label{section_intro}
\input{section1}

\section{Problem setup}\label{section_problem}
\input{section2}

%
%
%

\section{KRnet}\label{section_invertible}
\input{section3}

\input{section4}

\section{Adaptive deep density aproximation for the stationary Fokker-Planck equation}\label{section_fpequation}
\input{section5}

\section{Numerical study}\label{section_test}
\input{section6}

\section{Conclusions}\label{section_conclude}
Conducting adaptivity is of fundamental importance for the efficient approximation of high-dimensional Fokker-Planck equations.
With a focus on deep learning methods, we have developed an adaptive deep density approximation strategy based on KRnet (ADDA-KR) in this work. 
Our KRnet, which is built on a block-triangular structure inspired by the Knothe-Rosenblatt rearrangement,  gives an explicit family of probability density functions, which can serve as solution candidates of the Fokker-Planck equation.
We also showed that KRnet is effective for estimating high-dimensional density functions in general. 
\xw{The fact that KRnet can efficiently generate samples integrates the two main steps in 
our ADDA-KR strategy to achieve efficient iterations: train KRnet for the Fokker-Planck equation with current collocation
points, and generate new collocation points using the  KRnet for the next iteration.}
Compared to real NVP, which is a widely used generative model, numerical results show that our ADDA-KR gives much more accurate numerical solutions for the Fokker-Planck equation. ADDA-KR in general works very well for Fokker-Planck equations with dimension of $\mathit{O}(10)$. \xw{For higher-dimensional cases, the sparsity of high-dimensional data will induce more severe difficulties, where we may need to  consider dimension reduction to adapt more problem properties into the algorithm. }

\bigskip
\textbf{Acknowledgments:}
K. Tang and Q. Liao are supported by the National Natural Science Foundation of China (No. 12071291)
and the Science and Technology Commission of Shanghai Municipality (No. 20JC1414300), 
and X. Wan’s work was supported by the National Science Foundation under grant DMS-1913163.


\bibliography{tang}

\end{document}

%% file: section1.tex
During the past few decades there has been a rapid development in numerical methods for Fokker-Planck equations.
This explosion in interest has been driven by the need of assessing time evolution of probability density functions 
in randomly perturbed dynamical systems, which are widely used in physical and biological modeling \cite{risken1984fokker,ShiJin2011,YaoLi2019}. 
It is known that there exist two main challenges for efficiently solving the Fokker-Planck equations:
the spatial variable can be high-dimensional, which causes difficulties in applying grid based numerical methods,
e.g.\ finite element methods \cite{spencer1993numerical, elman2014finite};
the original spatial domain is typically unbounded, and it is challenging to derive a well-posed boundary condition for 
a bounded computational domain. 
To alleviate these difficulties, new numerical methods based on deep learning currently gain a lot of 
attention \cite{dlz2019efficientfp,ChenKarn2020}, and this paper is devoted to deep learning for the Fokker-Planck equations.

Deep learning methods for partial differential equations (PDEs) are under active development. 
In \cite{weinan2017proposal,weinan2018deep},
a deep Ritz method is proposed based on variational methods.
In \cite{raissi2017physics1,raissi2017physics2,raissi2019physics,pang2019fractional},
physics-informed neural networks are developed through infusing  PDEs into networks as a constraint. 
A deep Galerkin method is proposed in \cite{sirignano2018dgm}.
Bayesian deep convolutional encoder-decoder networks for PDEs with high-dimensional random inputs
are developed in \cite{zhu2018bayesian,zhu2019physics}. 
Deep learning strategies are also introduced to discover physical laws \cite{wu2019numerical,wu2019structure}. 
In addition, efficient deep learning methods based on domain decomposition are studied in \cite{li2019d3m,JagtKarn20,dong2020local,deepdd,kups20708,KharKarn2021},
and  deep neural network methods for complex geometries and irregular domains are proposed in \cite{sheng2020pfnn, WangGao21}.
The main idea of deep learning methods for PDEs 
is to reformulate a PDE problem as an optimization problem and train deep neural networks through minimizing the 
corresponding loss functional. 
In these methods, stochastic collocation points are required to estimate the loss functional.
We note that the stochastic collocation points herein are for the spatial variable, while stochastic collocation for 
PDEs with random inputs (especially for parameters) are discussed in detail in 
\cite{xiu2010numerical,Xiu2005High,babuvska2007stochastic,FooWan08,maza2009adaptive,xiu-nar12,lei2015constructing}. 
To result in an efficient deep learning strategy  for PDEs, properly choosing the collocation points is 
crucial. \xw{Intuitively, the distribution of the collocation points should be consistent with the properties of the PDE solution in a certain sense.}
In our recent work \cite{li2019d3m}, a hierarchical sampling procedure are proposed based on domain decomposition iterations, 
while it focuses on low-dimensional problems.
As the spatial variable of the Fokker-Planck equation can be high-dimensional, 
it remains an open challenging problem to generate effective collocation points.
We develop an effective adaptive sampling procedure to alleviate this issue in this work. \xw{Adaptivity is widely used in machine learning techniques to make the training process more effective by exploring the relation between the model and the data, e.g., active learning selects the most helpful samples to increase efficiency \cite{ren2020,CangCAD19} and meta-learning tries to match learning algorithms with task properties \cite{Vilalta2002}. In our problem, we will update the training set partially or completely according to the learned model, i.e., the approximate solution of the F-P equation, and the updated training set will yield a better approximate solution.}

As the solution of the Fokker-Planck equation is a probability density function,
solving this problem can also be considered as a density estimation problem.
It is known that density estimation is a central topic in unsupervised learning, and it still remains an open challenge 
for high-dimensional density estimation  \cite{scott2015multivariate}. Recently, two kinds of deep learning models 
have shown great promise for estimating high-dimensional probability density functions (PDFs), 
which include the flow-based generative model \cite{dinh2016density,kingma2018glow} and the 
neural ordinary differential equation model \cite{zhang2018monge,chen2018neural}.
In this work, we focus on the flow-based generative model, which is to construct invertible mappings from a prescribed prior distribution to the empirical distribution given by data and build explicit probability density functions using the change of variables. 
\xw{The Knothe-Rosenblatt (KR) rearrangement \cite{carlier2010knothe} shows that such an invertible mapping can be achieved with a triangular structure.
Incorporating with the KR rearrangement, we propose an invertible block-triangular mapping, called KRnet, 
which generalizes the flow-based generative model given by real NVP \cite{dinh2016density}.} 
We note that there are a lot of generative models which can efficiently generate samples of the distributions under consideration but 
do not explicitly give the corresponding density functions, e.g.,  generative adversarial networks (GANs) 
\cite{goodfellow2014generative} and the variational autoencoder (VAE) \cite{kingma2014auto}. In addition, coupling flow-based generative models and reduced-order models into an importance sampling estimator
is studied in \cite{wan2020coupling}.

In this work, 
we propose an adaptive deep density approximation method based on KRnet (ADDA-KR)
for solving Fokker-Planck equations. 
\xw{We first provide additional details and results for KRnet that was outlined in the letter \cite{tangwandensity2020}}.
After that, we use KRnet to construct solutions of the Fokker-Planck equation. 
Since KRnet can induce a family of probability density functions, 
normality and vanishing boundary conditions are satisfied naturally. 
\revs{Like other deep learning algorithms for solving PDEs, our method is also meshfree. The PDE problem is converted into an optimization problem and it can be solved through stochastic gradient descent on a set of collocation points, while traditional grid-based numerical methods (e.g. finite element methods) rapidly become computationally infeasible since the number of grid points grows exponentially with the dimensionality. The choice of the collocation points plays a crucial role in a meshless method. The distribution of the collocation points should be consistent with the regularity of the solution for both accuracy and efficiency. Since the solution of the F-P equation is a probability density function, one way to achieve this is to use the samples of the solution PDF as the collocation points.} 
Based on such an idea, we propose an adaptive approach ADDA-KR that has two main steps: 
training a KRnet to approximate the solution of the Fokker-Planck equation, 
and using the trained KRnet to generate collocation points for the next iteration. \xw{After each iteration, the distribution of the collocation points is more consistent with the solution PDF.}

The rest of the paper is organized as follows. In the next section, the Fokker-Planck equations and the problem setting are introduced. Our KRnet is presented in section \ref{section_invertible}. In section \ref{section_fpequation}, our novel adaptive deep density approximation  approach for the Fokker-Planck equation is presented. In section \ref{section_test}, we demonstrate the efficiency of our adaptive sampling approach with numerical experiments. Finally section \ref{section_conclude} concludes the paper.

%% file: section2.tex
Consider the state $X_t$ modeled by the following stochastic differential equation
\begin{equation} \label{eqn_fpst_ode}
dX_t = \bmu(X_t,t) dt + \bG(X_t,t) d \mb{w}_t,
\end{equation}
where $\bmu = [\mu_1, \ldots, \mu_d]^{\mathsf{T}}$ is a vector field, 
$\bG(X_t, t) \in \mathbb{R}^{d \times d}$ is a matrix-valued function 
and $\mb{w}_t$ is a $d$-dimensional standard Wiener process.
The Fokker-Planck equation, which describes the probability density function of $X_t$, 
is 
\begin{equation} \label{eqn_fp}
\begin{aligned}
\frac{\partial p(\mx,t)}{\partial t} = \mathcal{L} p(\mx, t) := \nabla \cdot \left [p(\mx,t) \nabla V(\mx,t) \right ] &+ \nabla \cdot [\nabla \cdot (p(\mx,t) \mb{D}(\mx, t))], \qquad \forall (\mx, t) \in \xs{R}^d \times \xs{R}^{+}, \\
\int_{\xs{R}^d } p(\mb{x},t) d \mx &= 1, \ p(\mx,t) \geq 0, \qquad \forall (\mx, t) \in \xs{R}^d \times \xs{R}^{+}, \\
\revs{p(\mb{x}, 0)} &= \revs{p_0(\mb{x}) },
\end{aligned}
\end{equation}
where $\mx \in \mathbb{R}^d$ denotes a random vector, $V(\mb{x},t)$ is a potential function, $\mb{D}(\mb{x},t)$ is a diffusion matrix,  $p(\mb{x},t)$ is the unknown probability density function (PDF) of $\mx$ \revs{with the initial PDF $p_0(\mb{x})$}, and $\mathcal{L}$ denotes the partial differential operator. 
Following \cite{risken1984fokker}, 
the potential function $V(\mb{x},t)$ and the diffusion matrix $\mb{D} (\mb{x},t)$ can be expressed as  
\begin{equation*}
\begin{aligned}
\nabla V(\mb{x},t) &= -\bmu(\mb{x},t),  \\
\mb{D}(\mb{x}, t)) &= \frac{1}{2}\bG(\mb{x},t) \bG(\mb{x},t)^\mathsf{T}.  
\end{aligned}
\end{equation*}
In this work, we focus on the stationary solution of Eq.\eqref{eqn_fp}, i.e., the invariant measure 
independent of time,
\begin{equation} \label{eqn_fpsta}
 \mathcal{L} p(\mb{x}) = \nabla \cdot \left [p(\mb{x}) \nabla V(\mb{x}) \right ] + \nabla \cdot [\nabla \cdot (p(\mb{x}) \mb{D}(\mb{x}))] = 0,
\end{equation}
with the boundary condition
\begin{equation} \label{eq_fpconst}
p(\mb{x}) \rightarrow 0 \quad \text{as} \quad \norm{\mb{x}}{2} \rightarrow \infty,
\end{equation}
and some extra constraints on $p(\bx)$ 
\begin{equation}\label{eq_pdf_positivity}
\int_{\xs{R}^d } p(\mb{x}) d \mb{x} = 1, \quad\textrm{and} \quad p(\mb{x}) \geq 0,
\end{equation}
where $\norm{\mb{x}}{2}$ indicates the $\ell_2$ norm of $\mb{x}$.

There are several difficulties for the approximation of equation \eqref{eqn_fpsta}. 
First, the boundary condition and the constraints 
of $p(\mb{x})$ may not be easily satisfied 
when we employ the traditional approaches such as the finite element method. 
Since the support of $p(\mb{x})$ is $\xs{R}^d$, the computation domain has to be truncated, implying that the boundary condition must be approximated, e.g., a homogeneous boundary condition. To preserve the nonnegativity of $p(\mb{x})$, a projection step is needed for the box constraint. Second, it requires a fine mesh to capture the whole information when the target density is multimodal, i.e., the potential function $V(\mb{x})$ has many local minima \cite{pavliotis2014stochastic}, which is computationally infeasible when the dimension $d$ is even moderately large. 
We also note that a homogeneous boundary condition usually requires a large computational domain, which makes a uniform refinement even more challenging, if no prior information can be used for certain adaptivity on mesh generation. 
To address these issues, we will propose an adaptive deep density approximation method to solve the Fokker-Planck equation \eqref{eqn_fpsta} using a deep generative model for $p(\mb{x})$. The flow-based generative model not only provides an explicit density function that satisfies naturally all constraints on $p(\mb{x})$, but also suggests a simple but effective adaptive strategy for the approximation of equation \eqref{eqn_fpsta} through sampling the current approximation of $p(\mb{x})$.

%% file: section3.tex
\xw{KRnet is a flow-based generative model for density estimation or approximation. In this section we briefly overview KRnet that has been outlined in our recently published letter \cite{tangwandensity2020} and present more details that were not included in \cite{tangwandensity2020} due to the page limit.} Let $X \in \xs{R}^d$ be a random vector associated with a given data set, and its  probability density function (PDF) 
is denoted by $p_X(\mb{x})$. \xw{The target is to estimate $p_X(\bx)$ using available data.} Let $Z \in \xs{R}^{d}$ be a random vector associated with a PDF $p_Z(\mb{z})$, where $p_Z(\mb{z})$ is a prior distribution (e.g., Gaussian distribution). The \comm{goal of} flow-based generative modeling is to seek an invertible mapping $\mb{z} = f(\mb{x})$ where $f(\cdot)$ is a bijection: $f:\mb{x} \mapsto \mb{z}$ \cite{dinh2016density}. By the change of variables, we have the PDF of $X=f^{-1}(Z)$ as 
\begin{equation}\label{eqn:pdf_model}
p_{X}(\mb{x})=p_{Z}(f(\mb{x})) \left |\det\nabla_{\mb{x}} f \right|.
\end{equation}
Once the prior distribution $p_{Z}(\mb{z})$ is specified, equation \eqref{eqn:pdf_model} provides an explicit PDF of $X$. \xw{Given a set of training data, the invertible mapping $f(\cdot)$ can be learned by maximizing the likelihood or minimizing the cross entropy. The inverse of $f(\cdot)$ provides a convenient way to sample $X$ as $X=f^{-1}(Z)$. }

\subsection{A new affine coupling layer} \label{sec_coupling_layer}
In flow-based generative models, the invertible mapping $f(\cdot)$ is constructed by stacking a sequence of simple bijections, each of which is a shallow neural network, and thus the overall mapping is a deep net. The mapping $f(\cdot)$ can be written in a composite form:
\begin{equation}\label{eqn:multi-layer-mapping}
\mb{z} = f(\mb{x}) = f_{[L]} \circ \ldots \circ f_{[1]}(\mb{x})\quad\textrm{ and }\quad\mb{x} = f^{-1}(\mb{z}) = f_{[1]}^{-1} \circ \ldots \circ f_{[L]}^{-1}(\mb{z}),
\end{equation}
where $f_{[i]}$ is called an affine coupling layer at stage $i$.  The Jacobian matrix can be obtained by the chain rule
\begin{equation}\label{eqn_f_det}
\left|\det \nabla_{\mb{x}}f\right| = \prod_{i=1}^L \left|\det\nabla_{\mb{x}_{[i-1]}}f_{[i]}\right|,
\end{equation}
where $\mb{x}_{[i-1]}$ indicate the intermediate variables with $\mb{x}_{[0]} = \mb{x}$ and $\mb{x}_{[L]} = \mb{z}$. 
Let $\mb{x}_{[i]} = [\mb{x}_{[i], 1},\mb{x}_{[i], 2}]^\mathsf{T}$ be a partition of $\mb{x}_{[i]}$ with $\mb{x}_{[i],1} \in \mathbb{R}^m$ and $\mb{x}_{[i],2} \in \mathbb{R}^{d-m}$ for $i = 0, \ldots, L-1$. \xw{One technique to define the affine coupling layer is the real NVP \cite{dinh2016density}:}
\begin{equation} \label{eqn_new_affine_coupling_rnvp}
\begin{aligned}
\mb{x}_{[i], 1} &= \mb{x}_{[i-1], 1} \\
\mb{x}_{[i], 2} &= \mb{x}_{[i-1], 2} \odot \exp\left(\log\mb{s}_i(\mb{x}_{[i-1], 1})\right)  + \mb{t}_i(\mb{x}_{[i-1], 1}),
\end{aligned}
\end{equation}
where $\mb{s}_i: \xs{R}^m \mapsto \xs{R}^{d-m}$ and $\mb{t}_i: \xs{R}^{m} \mapsto \xs{R}^{d-m}$ are the scaling and the translation depending on $\mb{x}_{[i-1], 1}$, and $\odot$ is the Hadamard product or element-wise product. \xw{Note that $\mb{x}_{[i-1],1}$ remains fixed and the modification of $\bx_{[i-1],2}$ is linear with respect to $\bx_{[i-1],2}$ and nonlinear in terms of $\bx_{[i-1],1}$. This way, the Jacobian matrix $\nabla_{\mb{x}_{[i-1]}}f_{[i]}$ is lower-triangular whose determinant can be evaluated efficiently. Furthermore, $(\mb{s}_i,\mb{t}_i)$ is usually modeled by a neural network $\mathsf{NN}_{[i]}$}
\begin{equation} \label{eq_nn_affine}
(\mb{s}_i, \mb{t}_i) = \mathsf{NN}_{[i]}(\mb{x}_{[i-1], 1}).
\end{equation}
 We proposed a new affine coupling layer $f_{[i]}$ as follows \cite{tangwandensity2020} 
\begin{equation} \label{eqn_new_affine_coupling}
\begin{aligned}
\mb{x}_{[i], 1} &= \mb{x}_{[i-1], 1} \\
\mb{x}_{[i], 2} &= \mb{x}_{[i-1], 2} \odot \left(1 + \alpha \ \mathrm{tanh}(\mb{s}_i(\mb{x}_{[i-1], 1})) \right) + e^{\mb{\beta}_i} \odot \mathrm{tanh}(\mb{t}_i(\mb{x}_{[i-1], 1})),
\end{aligned}
\end{equation}
\xw{where $0<\alpha<1$ is a hyperparameter and the parameter $\mb{\beta}_i \in \xs{R}^{d-m}$ is trainable. Our affine coupling layer keeps the mechanism of the real NVP when updating the data, and it has the following advantages.} First, the second equation in Eq.\eqref{eqn_new_affine_coupling} adapts the trick of ResNet \cite{he2016resnet}, where an identity mapping is added to improve the training process. Second, the constant $\alpha \in (0,1)$ is introduced to improve numerical stability. 
It is seen that the range of \revs{$\det \nabla_{\mb{x}_{[i-1]}} {f_{[i]}}$} is 
$[(1 - \alpha)^{d-m}, (1 + \alpha)^{d-m}]$ for our affine coupling layer and $(0,+\infty)$ for the original real NVP.  
Our formulation can alleviate the illnesses when the determinant of the Jacobian in the original real NVP occasionally become too large or too small. \revs{Third, the trainable factor $e^{\mb{\beta}_i}$ depends on the whole training set, which helps avoid possible large oscillation in $\mb{t}_{i}(\mb{x}_{[i-1],1})$ such that the number of outliers can be reduced for sample generation \cite{tangwandensity2020}. In our numerical experiments, we set $\alpha = 0.6$ and it works well.}

Since the affine coupling layer $f_{[i]}$ only updates a part of $\mb{x}_{[i-1]}$, another affine coupling layer is needed for a complete update. In other words, the next affine coupling layer $f_{[i+1]}$ can be defined as
\begin{equation*}
\begin{aligned}
\mb{x}_{[i+1], 1} &= \mb{x}_{[i],1} \odot \left( 1 + \alpha \ \mathrm{tanh}(\mb{s}_{i+1}(\mb{x}_{[i],2})) \right) + e^{\mb{\beta}_{i+1}} \odot \mathrm{tanh} \left( \mb{t}_{i+1}(\mb{x}_{[i],2}) \right) \\
\mb{x}_{[i+1],2} &= \mb{x}_{[i],2},
\end{aligned}
\end{equation*}
where the components $\mb{x}_{[i],1}$ are updated and $\mb{x}_{[i],2}$ remains unchanged. \xw{From the dynamical point of view, a long chain of affine coupling layers may result in a highly nonlinear transformation of the input. To enhance the performance and efficiency of the mapping $f(\mb{x})$, we proposed KRnet to address the following questions: 1) How should we partition the vector? 2) How can we increase the modeling capability except for increasing the depth $L$? 3) Can we provide a robust nonlinear bijection at least in a component-wise way?}

%% file: section4.tex
\subsection{The overall structure of KRnet}
\xw{The basic idea of KRnet is to define the structure of $f(\mb{x})$ in terms of the Knothe-Rosenblatt rearrangement. }
Let $\mu_Z$ and $\mu_X$ be the probability measures of two random variables $X,Z\in\xs{R}^d$ respectively. A mapping $\mathcal{T}$: $Z \mapsto X$ is called a transport map such that $\mathcal{T}_{\#} \mu_Z = \mu_X$, where $\mathcal{T}_{\#} \mu_{Z}$ is the push-forward of $\mu_{Z}$ such that $\mu_{X}(B) = \mu_{Z}(\mathcal{T}^{-1}(B))$ for every Borel set $B$ \cite{carlier2010knothe}. The Knothe-Rosenblatt rearrangement tells us that the transport map $\mathcal{T}$ may have a lower-triangular structure 
\begin{equation}
    \mb{z} = \mathcal{T}^{-1}(\mb{x}) = \left[ 
    \begin{array}{l}
    \mathcal{T}_1(x_1) \\
    \mathcal{T}_2(x_1, x_2) \\
    \vdots \\
    \mathcal{T}_{d}(x_1, \ldots, x_d)
    \end{array}
    \right].
\end{equation}
This mapping can be regarded as a limit of sequence of optimal transport maps when the quadratic cost degenerates \cite{carlier2010knothe}. \xw{Noticing that the invertible mapping $f(\bx)$ also defines a transport map, we then incorporate the triangular structure of the Knothe-Rosenblatt rearrangement into the definition of $f(\bx)$ which results in KRnet as a generalization of real NVP \cite{dinh2016density}}. Let $\mb{x} = [\mb{x}^{(1)}, \ldots, \mb{x}^{(K)}]^\mathsf{T}$ be a partition of $\mb{x}$, where $\mb{x}^{(i)} = [x_{1}^{(i)}, \ldots, x_{m}^{(i)}]^\mathsf{T}$ with $1 \leq K \leq d, 1 \leq m \leq d$, and $\sum_{i=1}^K \mathrm{dim}(\mb{x}^{(i)}) = d$. Our KRnet takes an overall form
\begin{equation} \label{eqn_KR}
  \mb{z} = f_{\mathsf{KR}}(\mb{x}) = \left[ 
    \begin{array}{l}
    f_1(\mb{x}^{(1)}) \\
    f_2(\mb{x}^{(1)}, \mb{x}^{(2)}) \\
    \vdots \\
    f_{K}(\mb{x}^{(1)}, \ldots, \mb{x}^{(K)})
    \end{array}
    \right],
\end{equation}
where each $f_{i}$ is an invertible mapping defined as in equation \eqref{eqn:multi-layer-mapping} for $ i = 2, \ldots, K$. \xw{Note that $f_1$ is not included if $K = d$ because we need to partition a vector to two parts to define the affine coupling layer.}  KRnet consists of one  outer loop and $K-1$ inner loops. The outer loop has $K-1$ stages, corresponding to the $K-1$ mappings $f_{i}$ in equation \eqref{eqn_KR} with $i=2,\ldots,K$, and for each stage, an inner loop of $L$ affine coupling layers is defined. More specifically, we have
\begin{equation}
\mb{z} = f_{\mathsf{KR}}(\mb{x}) = L_{N} \circ f_{[K-1]}^{\textsf{outer}} \circ \cdots \circ f_{[1]}^{\textsf{outer}} (\mb{x}),
\end{equation}
where $f_{[i]}^{\textsf{outer}}$ is defined as
\begin{equation}
f_{[k]}^{\textsf{outer}} = L_S \circ f_{[k, L]}^{\textsf{inner}} \circ \cdots \circ f_{[k,1]}^{\textsf{inner}} \circ L_R.
\end{equation}
Here $f_{[k,i]}^{\textsf{inner}}$ indicates a combination of one affine coupling layer and one scale and bias layer, and $L_N$, $L_S$ and $L_R$ indicate the nonlinear layer, the squeezing layer and the rotation layer, respectively, which will be briefly overviewed in the next section.

\begin{figure}
\centering
\begin{tikzpicture}
  [node distance=.5cm,
  start chain=going below,]
     \node[punktchain, join] (input) {$\mb{x}$};
     \node[punktchain, join] (rotation)      {$L_R$: Rotation layer};
     \node[punktchain, join] (scale)      {Scale and bias layer};
     \node[punktchain, join] (affine) {Affine coupling layer};
     \node[punktchain, join, ] (squeeze) {$L_S$: Squeezing layer};
     \node[punktchain, join, ] (nonlinear) {$L_N$: Nonlinear layer for the rest (or all) dimensions};
     \node[punktchain, join, ] (output) {$\mb{z}$};
  \draw[tuborg, decoration={brace}] let \p1=(scale.north), \p2=(affine.south) in
    ($(2, \y1)$) -- ($(2, \y2)$) node[tubnode] {$f_{[k,i]}^{\textsf{inner}}$};
  \draw[thick, ->]  
      (affine.west)
      -- ++ (-1.5cm, 0) 
      node[pos=0.65,above,sloped] {$L$} |-  (scale.west);
   \draw[ultra thick, ->]  
          (squeeze.west)
          -- ++ (-2.5cm, 0) 
          node[pos=0.65,above,sloped] {$K-1$} |-  (rotation.west);  
\end{tikzpicture}
\caption{The flow chart of KRnet.}
\label{fig_krnet}
\end{figure}
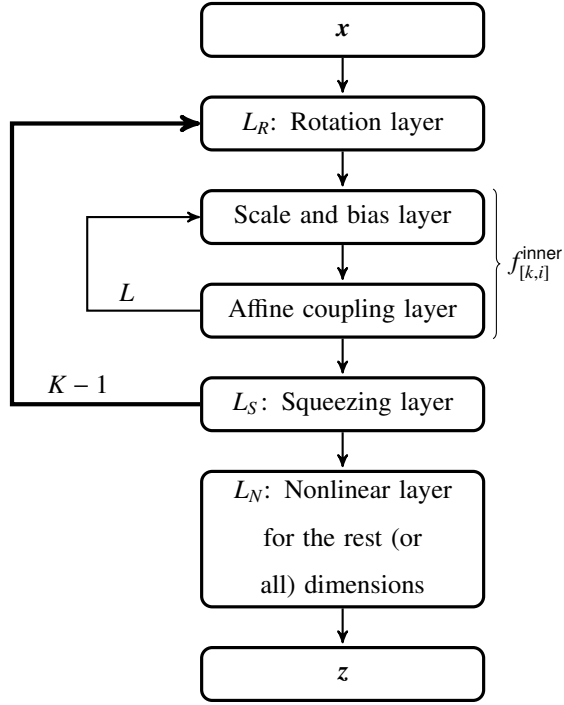

The flow chart of KRnet is illustrated in Figure \ref{fig_krnet}. Let us look at how the information flows in the KRnet. Each $\mb{x}_{[k]} = [\mb{x}_{[k]}^{(1)}, \ldots, \mb{x}_{[k]}^{(K)}]^\mathsf{T}$ has the same partition with $\mb{x}_{[k]} = f_{[k]}^{\textsf{outer}}(\mb{x}_{[k-1]})$ with $\mb{x}_{[0]}=\mb{x}$, $k=1,\ldots,K-1$. 
At the beginning, a sequence of affine coupling layers in $f_{[1]}^{\textsf{outer}}$ is applied to the partition $\mb{x}_{[0]}=[\mb{x}_{[0]}^{(1:K-1)},\mb{x}_{[0]}^{(K)}]^\mathsf{T}$, where $\mb{x}_{[0]}^{(1:K-1)}$ includes $\mb{x}_{[0]}^{(i)}$, $i=1,\ldots,K-1$. From then on, the last partition $\mb{x}_{[k]}^{(K)}$ will remain fixed for $k>1$. For the next iteration $f_{[2]}^{\textsf{outer}}$, the partition $[\mb{x}_{[1]}^{(1:K-2)},\mb{x}_{[1]}^{(K-1)}]^\mathsf{T}$ will be used with $\mb{x}_{[1]}^{(K)}$ being deactivated. In general, after the stage $K - i + 1$ of the outer loop, the $i$-th partition of $\mb{x}_{[k]}^{(i)}$ will become deactivated, in addition to the dimensions that are deactivated in the previous stages. 

\comm{The inner loop is composed of a sequence of general coupling layers $f_{[k, i]}^{\textsf{inner}}$
\begin{equation}
f_{[k]}^{\textsf{outer}} = L_S \circ f_{[k, L]}^{\textsf{inner}} \circ \cdots \circ f_{[k,1]}^{\textsf{inner}} \circ L_R,
\end{equation}
where $L_S$ is a squeezing layer and $L_R$ is a rotation layer. Each general coupling layer $f_{[k,i]}^{\textsf{inner}}$ includes a scale and bias layer and $L$ affine coupling layer as discussed in section \ref{section_flow}. The main difference here is to introduce a new nonlinear layer in the general coupling layer, which will be discussed in section \ref{sec:KRnet_layers}. The flow chart of KR net can be illustrated in Figure \ref{fig_krnet}. The data first go through a rotation layer to reduce the correlation between each dimension through a linear combination. Then $L$ general coupling layers are defined, after which a certain portion of the active components will become inactive after a squeezing layer. This filtering process will be repeated $K-1$ times to achieve the overall invertible mapping.}

\subsection{Other types of layer used in KRnet}\label{sec:KRnet_layers}
\xw{Except for the affine coupling layers, several other types of layers are needed for the definition of KRnet. We briefly overview these layers in this section and provide some details excluded in the letter \cite{tangwandensity2020}. Since each $\mb{x}_{[k]}$ has the same partition, we will drop the subscript for simplicity.}

\emph{Squeezing layer $L_S$} is used to deactivate some dimensions using a mask
\begin{equation}\label{eqn_squeeze}
    \mb{q} = [\underbrace {{1,\ldots,1} }_{n}, \underbrace {{0,\ldots,0 }}_{d-n}]^\mathsf{T},
\end{equation}
where the components $\mb{q}\odot\mb{x}$ will keep being updated and the rest components $(1-\bq)\odot\mb{x}$ will be fixed from then on. 

\xw{\emph{Scale and bias layer} provides a simplification of the batch normalization \cite{ioffe2015batch}, which is defined as  
\begin{equation}\label{eqn:actnorm}
	\hat{\mb{x}} = \mb{a} \odot \mb{x} + \mb{b},
\end{equation}
where $\mb{a}$ and $\mb{b}$ are trainable and initialized by the mean and standard deviation of the data. After the initialization, $\mb{a}$ and $\mb{b}$ will be treated as regular trainable parameters that are independent of the data. Numerical experiments show that the scale and bias layer is simple but effective, which provides a comparable performance to the batch normalization layer in our problem setting. }

\emph{Rotation layer $L_R$} defines a linear mapping of the input $\mb{x}$ 
\begin{equation*}
    \hat{\mb{x}} = \hat{\mb{W}} \mb{x}.
\end{equation*}
through a trainable matrix \comm{is defined through an invertible matrix that is learnable}
\begin{equation*}
    \hat{\mb{W}} = \left[ \begin{array}{cc}
         \mb{W} & \mb{0}  \\
         \mb{0} & \mb{I}
    \end{array}
    \right]  = \left[ \begin{array}{cc}
	\mb{L} & \mb{0}  \\
	\mb{0} & \mb{I}
\end{array}
\right] \left[ \begin{array}{cc}
	\mb{U} & \mb{0}  \\
	\mb{0} & \mb{I}
\end{array}
\right],
\end{equation*}
where $\mb{W} \in \xs{R}^{n \times n}$, $n$ is defined in the mask $\mb{q}$, $\mb{I} \in \xs{R}^{(d-n) \times (d-n)}$ is an identity matrix, and $\mb{W}=\mb{L}\mb{U}$ is the LU factorization of $\mb{W}$. \xw{We expect $\hat{\mb{W}}$ to provide a rotation such that the less important dimensions will be put at the end and deactivated by the next squeezing layer. Entries below the main diagonal of $\mb{L}$ and entries in the upper triangle of $\mb{U}$ are trainable. In practice, we simply optimize the trainable entries of $\mb{L}$ and $\mb{U}$ without  enforcing the orthonormality of $\hat{\mb{W}}$ and such a simplification works well.}

\emph{Nonlinear layer $L_N$} provides a component-wise nonlinear transformation. For simplicity, we only consider one component $x$ of the data. We start with a nonlinear mapping  $F(s): [0,1] \mapsto [0,1]$:
\begin{equation}\label{eqn_nonlinear_layer}
    F(s) = \int_{0}^{s} p(t) dt,
\end{equation}
where $p(s)$ is a probability density function. Let $0 = s_0 < s_1 < \ldots < s_{\hat{m}+1} = 1$ be a mesh of the interval $[0,1]$ with element size $h_i=s_{i+1}-s_i$. Define $p(s)$ as a piece-wise linear polynomial 
\begin{equation}
p(s)=\frac{w_{i+1}-w_i}{h_i}(s-s_i)+w_i,\quad\forall s\in[s_i,s_{i+1}],
\end{equation}
where
\[
p(s_i)=w_i.
\]
Then $F(s)$, corresponding to a cumulative density function, is a quadratic function
\begin{equation}
F(s)=\frac{w_{i+1}-w_i}{2h_i}(s-s_i)^2+w_i(s-s_i)+\sum_{k=0}^{i-1}\frac{w_k+w_{k+1}}{2}h_i,\quad\forall s\in[s_i,s_{i+1}],
\end{equation}
whose inverse and derivative can be explicitly computed. 

As the support of each dimension of $\mb{x}$ is $(-\infty, \infty)$, a question is how to apply $F(s)$ to the data. A straightforward strategy is to map $(-\infty, \infty)$ to $(0,1)$ before $F(s)$ is applied. However, when the inverse is considered, the singularity of mapping a finite interval to an infinite one may introduce issues on robustness. To alleviate this problem, we decompose $(-\infty, \infty) = (-\infty, -a) \cup [-a, a] \cup (a, \infty)$ with $a > 0$, 
and define the following nonlinear mapping
\begin{equation}
\hat{F}(x)=\left\{
\begin{array}{rl}
\beta_s (x+a)-a, &x\in(-\infty,-a)\\
2aF\left(\frac{x+a}{2a}\right)-a,&x\in[-a,a]\\
\beta_s (x-a)+a, &x\in(a,\infty),
\end{array}
\right.
\end{equation}
where $\beta_s>0$ is a scaling factor. It is seen that we only consider a nonlinear mapping for the data located in $[-a,a]$ and $\hat{F}(x)$ maps $[-a,a]$ to itself. On $(-\infty,-a)\cup(a,\infty)$, $\hat{F}(x)$ is simply a linear mapping. The reasoning of such a strategy is that the range of data in the training set is always finite, and after being well scaled and shifted the data will be roughly centered at the origin, implying that a nonlinear mapping on $[-a,a]$ is sufficient as long as $a$ is large enough. To maintain the invertibility, we require some regularity at $x=\pm a$. More specifically, $\hat{F}'(x)$ should exist at $x=\pm a$. Since $\hat{F}'(x)=\beta_s$ on $(-\infty,-a)\cup(a,\infty)$, we have, on $[-a,a]$, $\hat{F}'(x)|_{x=\pm a}=F'(s)|_{s=0,1}=p(s)|_{s=0,1}=\beta_s$. So the trainable parameters include $p(s_i)=w_i$, $i=1,\ldots,\hat{m}$, subject to the constraint $\int_0^1p(s)ds=1$. 

\begin{rmk}
\xw{The nonlinear layer is only employed before the final output (see Figure 
\ref{fig_krnet}), which can be applied to all dimensions or simply the dimensions that have not been deactivated by the squeezing layer. In both cases, the nonlinear layer enlarges the prescribed prior distribution by a nonlinear component-wise transformation.}
The parameter $\beta_s$ acts as an estimate of the density $p(s)$ at $s=0,1$. If $a$ is sufficiently large, $\beta_s$ can be small accordingly. The prior distribution is often chosen as the standard Gaussian, which means that the density is larger around the origin when the data pass the nonlinear layer. This suggests we may consider an adaptive mesh for more effectiveness, in other words, the mesh is finer around $s=1/2$ and coarser around $s=0,1$.
\end{rmk}
 
\subsection{The complexity of KRnet}
We count the number of trainable parameters in KRnet. For simplicity, we assume that each $f_{[k]}^{\textsf{outer}}$ has $L$ general coupling layers $f_{[k,i]}^{\textsf{inner}}$. Let $d_k$ be the number of effective dimensions for $f_{[k]}^{\textsf{outer}}$ and $N_{\textsf{NN},k}$ the number of model parameters for the neural network Eq.\eqref{eq_nn_affine} used in $f_{[k,i]}^{\textsf{inner}}$. We note that the main characetristic of KRnet is that a portion of dimensions will be deactivated as $k$ increases. As $d_k$ decreases with $k$, we expect that the neural network Eq.\eqref{eq_nn_affine} in $f_{[k,i]}^{\textsf{inner}}$ should become simpler for a larger $k$. In other words, $N_{\textsf{NN},k}$ may decrease as $k$ increases. For simplicity, we let $N_{\textsf{NN},k}=rN_{\textsf{NN},k-1}$, where $0<r<1$, without worrying about the detailed configuration of the neural network. The number of trainable parameters is $d_k^2$ for $L_R$, and $\hat{m}d$ for $L_N$, and $2d_k$ for the scale and bias layer. Assume that $d=mK$. We have $d_k=d-(k-1)m$, $k=1,\ldots,K$. According to the flow chart in Figure \ref{fig_krnet}, we have the total number of model parameters as
\begin{equation}
N_{\mathsf{dof}}=\hat{m}d+\sum_{k=1}^{K-1}(N_{\textsf{NN},1}r^{k-1}L+(K-k+1)^2m^2+2(K-k+1)mL).
\end{equation}
The model complexity is mainly determined by the depth $L$ and the number $K$ for the partition of data. 

\subsection{KRnet for density estimation}\label{sec_KRnet_density}
\xw{We study the performance of KRnet for density estimation in this part and provide more results on the comparison between the real NVP and the KRnet that were not included in \cite{tangwandensity2020}.   Once the KRnet is constructed, we train the model  $p_X(\mb{x};\Theta)$ by maximizing the likelihood of the data or minimizing the cross entropy between the data distribution and the density model, where $\Theta$ includes all the trainable model parameters.} Let $\mathcal{S} = \{\mb{x}^{(i)}\}_{i=1}^{N_t}$ be the training set and $p_{X,\mathsf{data}}(\mb{x})$ the underlying data distribution. The Kullback-Leibler (KL) divergence between $p_{X,\mathsf{data}}(\mb{x})$ and  $p_{X}(\mb{x};\Theta)$ is
\begin{equation} \label{eq_kl_cross}
\underset{\Theta}{\mathrm{min}} \ D_{KL}(p_{X,\mathsf{data}}(\mb{x})|| p_{X}(\mb{x};\Theta)) = \mathbb{E}_{\mb{x} \sim p_{X,\mathsf{data}}(\mb{x})} \left[ \mathrm{log} \ \frac{p_{X,\mathsf{data}}(\mb{x})}{p_{X}(\mb{x};\Theta)} \right] = H\left( p_{X,\mathsf{data}}(\mb{x}), p_X(\mb{x};\Theta) \right) - H \left( p_{X,\mathsf{data}}(\mb{x}) \right)
\end{equation}
where $H \left( p_{X,\mathsf{data}}(\mb{x}) \right)$ is the entropy of $p_{X,\mathsf{data}}(\mb{x})$ and $H\left( p_{X,\mathsf{data}}(\mb{x}), p_X(\mb{x};\Theta) \right)$ is the cross entropy of $p_{X,\mathsf{data}}(\mb{x})$ and $p_X(\mb{x};\Theta)$. Since $p_{X,\mathsf{data}}(\mb{x})$ is independent of $\Theta$, minimizing the KL divergence is equivalent to minimizing the cross entropy. Note that
\begin{equation}
H(p_{X,\mathsf{data}}(\mb{x}),p_{X}(\mb{x};\Theta))\approx -\frac{1}{N_t}\sum_{i=1}^{N_t}\log p_{X}(\mb{x}^{(i)};\Theta),
\end{equation} 
which corresponds to the negation of the log-likelihood. 

To measure the quality of KRnet, we compute the KL divergence Eq.\eqref{eq_kl_cross} on a validation set between a reference PDF and the trained density model. The training data sets $\mathcal{S}$ is  generated as follows. Assume that $X$ has i.i.d. components and each component $X_i\sim\mathrm{Logistic}(0,s)$ has a PDF $\rho(x_i;0,s)$. We  generate a sample $\mb{x}^{(i)}$ of $X$, and then check if it satisfies the following constraint: 
\begin{equation}\label{eqn_logi_hole}
    \norm{\mb{R}_{\gamma, \theta_j} [x_j^{(i)}, x_{j+1}^{(i)}]^\mathsf{T}}{2} \geq C, \quad j = 1,\ldots,d-1,
\end{equation}
where $C$ is a specified constant, and 
\begin{equation*}
\mb{R}_{\gamma, \theta_j} = \left[ \begin{array}{cc}
         \gamma & 0  \\
         0 & 1
    \end{array}
    \right] \left[ \begin{array}{cc}
         \mathrm{cos} \theta_j & -\mathrm{sin} \theta_j  \\
         \mathrm{sin} \theta_j & \mathrm{cos} \theta_j
    \end{array}
    \right], \quad \theta_j = \left\{\begin{array}{ll}  \frac{\pi}{4}, & \ \text{if} \ j \ \text{is even}\\ 
                                                                             \frac{3\pi}{4}, & \ \text{otherwise}\end{array}\right..
\end{equation*}
The sample $\mb{x}^{(i)}$ will be accepted if the constraint Eq.\eqref{eqn_logi_hole} is satisifed and rejected otherwise. 
This way, an elliptic hole is generated for any two adjacent dimensions of data points. The reference PDF is then defined as
\begin{equation}
p_{X,\mathsf{ref}}(\mb{x})=\frac{I_B(\mb{x})\prod_{i=1}^d\rho(x_i;0,s)}{\mathbb{E}[I_{B}(X)]},
\end{equation}
where $B$ is the set defined by equation \eqref{eqn_logi_hole} and $I_B(\cdot)$ is an indicator function with $I_B(\mb{x})=1$ if $\mb{x}\in B$; 0, otherwise.  
For this test problem, we set $d = 8$, $\gamma =3$ and $C = 7.6$. 
This case has been studied in \cite{tangwandensity2020}, where the rotation layers and nonlinear layers are turned off. In \cite{tangwandensity2020} an algebraic convergence has been observed numerically for both the real NVP and the KRnet, where the convergence rate of KRnet is about twice as large as that of the real NVP. We here only demonstrate the effectiveness of the rotation layer and the nonlinear layer. 

We now compare the performance of KRnet and real NVP numerically. In KRnet, we deactivate the dimensions by one, i.e., $K=7$. We let $N_{\mathsf{NN},k}=0.9 N_{\mathsf{NN},k-1}$ by adjusting the width of the neural network $\mathsf{NN}_{[i]}$, $i=1,2,3$, which consists of two fully connected hidden layers of the same width. Other configurations of $\mathsf{NN}_{[i]}$ can also be considered. One example is given in Figure \ref{fig:nn_affine_coupling}, \revs{which is used in section \ref{section_test}}. The neural network $\mathsf{NN}_{[i]}$ (for $i = 0, \ldots, L-1$) consists of three hidden layers and one linear layer, where the first hidden layer and the linear layer have $w$ neurons, and the middle two layers have $w/2$ neurons. In this experiment, we combine the two middle hidden layers to one hidden layer with $w$ neurons. We set $w=24$ and use the rectified linear unit function (ReLU) as the activation function \cite{glorot2011deep}. The depth of the real NVP will be determined by $N_{\mathsf{dof}}$ of the KRnet, since we split the dimensions into two halves in real NVP. The KRnet will be implemented as follows. We train KRnet with three stages and record the errors of each stage. In the first stage, we switch off both the rotation layers and the nonlinear layers and train the model for 8000 epochs; in the second stage, we switch on the rotation layers and restart the training process for another 2000 epochs; finally, we switch on both the rotation layers and the nonlinear layers and continue the training process for another 2000 epochs. For the real NVP, we simply run 8000 epochs. For each epoch, we compute the relative error
\begin{equation} \label{eq_relative_error}
\delta=\frac{D_{KL}(p_{X,\mathsf{ref}}(\mb{x})|| p_{X}(\mb{x};\Theta))}{H(p_{X,\mathsf{ref}}(\mb{x}))} 
\end{equation}
using the validation set, since the cross entropy should converge to the differential entropy of the reference PDF. We record the minimum relative error of all epoches. Furthermore, to reduce the bias of $\delta$, we will sample 10 independent training sets and repeat the training process ten times to obtain an averaged relative error $\delta$. The relative errors corresponding to the above three stages of training KRnet are denoted as $\delta_I, \delta_{II}$ and $\delta_{III}$. We will sample $3.2\times 10^5$ data points for both the training set and the validation set. We employ the Adam optimizer \cite{kingma2014adam} with learning rate 0.001 and batch size 80000. 

\begin{figure}
	\center{
		\includegraphics[width=0.99\textwidth]{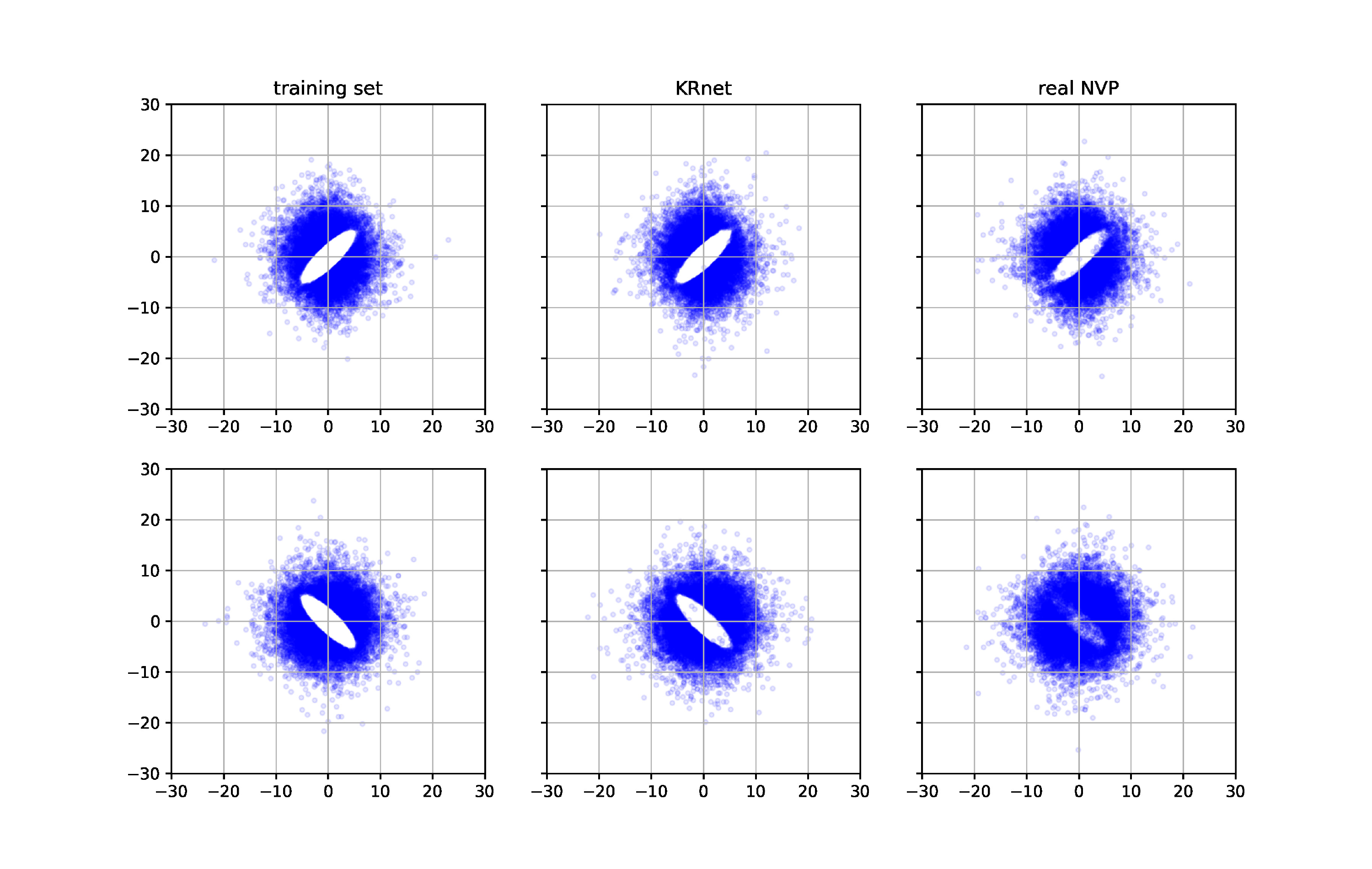}
	}
	\caption{Training data, and data sampled from KRnet and real NVP. The first row shows the components $x_1$ and $x_2$, and the second row shows the components $x_4$ and $x_5$. We pick the two pairs of adjacent dimensions, where the real NVP performs the best and the worst, respectively.}\label{fig:krnet_rNVP}
\end{figure}

The results of numerical experiments have been summarized in Table \ref{tab:krnet_r_n}. First of all, both $\delta_{II}$ and $\delta_{III}$ are smaller than $\delta_I$, indicating that the rotation layers and nonlinear layers are able to improve the model performance. Such an improvment is more noticeable for a smaller $L$. Second, for the specific setup of the numerical experiments, the errors $\delta_i$, $i=I,II,III$, of the KRnet decay consistently as $L$ increases while the errors of the real NVP do not show consistent decay. Since we compute the errors after 8000 epochs for all $L$, this shows that for a comparable model complexity the KRnet needs less epoches to obtain a substantial decrease in error than the real NVP. Third, as also shown in \cite{tangwandensity2020}, the real NVP performance better than KRnet for a small $L$. The real NVP can be regarded as a KRnet with a half-half partition, i.e., $K=2$ and $m=\frac{d}{2}=4$. For a fixed complexity, the performance of KRnet depends on both $K$ and $L$. In Figure \ref{fig:krnet_rNVP}, we compare the approximated distributions given by the real NVP with $L=42$ and the KRnet with $L=8$, where both the rotation layers and the nonlinear layers are switched on. 

\begin{table}[H]
    \caption{The effects of rotation and nonlinear layers in KRnet. $\delta_{I}$, $\delta_{II}$ and $\delta_{III}$ are relative errors of KRnet, respectively, for the aforementioned three stages. $\delta$ is the relative error of real NVP, whose depth is chosen to roughly match the DOFs of the KRnets from the same column. For the nonlinear layers, we use 32 nonuniform elements to decompose $[-30,30]$, i.e., $a=30$. Note that the rotation layers and the nonlinear layers do not introduce a significant increase in the total number of DOFs. The percentages in parentheses indicate the degree of drop in terms of $\delta_I$.}
    \label{tab:krnet_r_n}
    \begin{center}
        \begin{tabular}{|c|c|c|c|c|c|c|}\hline
            KRnet    & $L=2$  & $L=4$ & $L=6$ & $L=8$  \\ \hline
            $\delta_I$     & 7.54e-2 & 2.45e-2 & 1.44e-2 & 9.50e-3  \\ \hline
            $\delta_{II}$  & 6.53e-2 ($\downarrow$13\%) & 2.24e-2 ($\downarrow$9\%) & 1.39e-2 ($\downarrow$3\%) & 9.11e-3 ($\downarrow$4\%) \\ \hline
            $\delta_{III}$ & 4.93e-2 ($\downarrow$35\%) & 1.95e-2 ($\downarrow$20\%) & 1.26e-2 ($\downarrow$13\%) & 8.34e-3 ($\downarrow$12\%) \\ \hline
            Real NVP & $L=10$  & $L=20$ & $L=32$ & $L=42$   \\ \hline
            $\delta$ & 2.17e-2  & 1.98e-2 & 2.11e-2 & 2.05e-2  \\ \hline
        \end{tabular}
    \end{center}
\end{table}

\begin{figure}
	\begin{center}
		\smartdiagramset{back arrow disabled=true,module minimum height=2cm}
		\smartdiagram[flow diagram:horizontal]{
			input, FC layer with width $w$ + activation, FC layer with width $w/2$ + activation, FC layer with width $w/2$ + activation, FC layer with width $w$, output}
	\end{center}
	\caption{The architecture of $\mathsf{NN}_{[i]}$ for affine coupling layers, for $i = 0, \ldots, L-1$ (FC layers refer to fully connected layers).} \label{fig:nn_affine_coupling}
\end{figure}

%% file: section5.tex
We intend to use KRnet as a PDF model to approximate the Fokker-Planck equation to alleviate the difficulties from the curse of dimensionality. 
In particular, we will develop an adaptive deep density approximation (ADDA) approach, which consists of two components: 1) solving the Fokker-Planck equation on a certain set of collocation points by a machine learning technique; 2) choosing a new set of collocation points to \xw{refine the current approximate solution}. These two components are implemented alternately to achieve adaptivity such that both the accuracy and the efficiency will be improved. 

\subsection{Stochastic gradient descent based on stochastic collocation points} \label{sec_dlpde}
Let $p_{X}(\mb{x};\Theta)$ be a probability density function associated with the random vector $X$, which is based on the KRnet. 
All the constraints in Eq.\eqref{eq_fpconst} and Eq.\eqref{eq_pdf_positivity} are naturally satisfied since $p_X(\mb{x};\Theta)$ is a family of probability density functions, implying that the difficulties caused by the boundary conditions and the nonnegativity of PDF have disappeared. We seek to approximate the solution $p(\mb{x})$ of the Fokker-Planck equation by $p_{X}(\mb{x};\Theta)$ 
to take advantage of the weaker dependence of deep neural networks on dimensionality than traditional computational approaches such as the finite element methods \cite{cybenko1989approximation, leshno1993multilayer, lu2017expressive}. 

The main idea of a machine learning approach to solve PDEs is to consider an optimization problem defined on a set of collocation points where the equation is constrained. Let $p_{\mathsf{data}}(\mb{x})$ be a probability density function, based on which we define a loss functional
\begin{equation} \label{eq_resloss}
J \left( p_{X}(\mb{x};\Theta) \right)=\mathbb{E}_{p_{\mathsf{data}}(\mb{x})} \left( r^2(\mb{x};\Theta) \right) = \mathbb{E}_{p_{\mathsf{data}}(\mb{x})} \left( \mathcal{L}^2(p_X(\mb{x};\Theta)) \right)
\end{equation} 
where $\mathbb{E}_{p_{\mathsf{data}}(\mb{x})}$ denotes the expectation with respect to the training set, and $r$ is the residual loss. The solution $p(\mb{x})$ of Eq.\eqref{eqn_fpsta} can be approximated by $p_{X}(\mb{x};\Theta)$ through minimizing the loss functional $J(p_X(\mb{x};\Theta))$. In reality, we usually do not have much prior understanding about the residual, and simply assign $p_{\mathsf{data}}(\mb{x})$ a simple distribution, e.g., a uniform distribution defined on a finite computational domain. We then use $p_{\mathsf{data}}(\mb{x})$ to sample a set $\mathcal{C} = \{ \mb{x}^{(i)}\}_{i=1}^{N}$ of collocation points to approximate the loss functional, i.e.,
\begin{equation}\label{eqn_fploss}
 \hat{J} \left( p_{X}(\mb{x};\Theta) \right) = \frac{1}{N}\sum_{i=1}^N\mathcal{L}^2\left(p_{X}(\mb{x}^{(i)};\Theta)\right) \approx J \left( p_{X}(\mb{x};\Theta) \right),
\end{equation}
based on which we choose the optimal parameter $\Theta^*$: 
\begin{equation}\label{eqn:opt_prob}
\Theta^*=\argmin_{\Theta} \hat{J}(p_X(\mb{x};\Theta)).
\end{equation}

The optimization problem Eq.\eqref{eqn:opt_prob} will be solved by \revs{stochastic gradient-based optimization} \cite{bottou2018optimization, kingma2014adam}, which is summarized as follows. 
The set of collocation points can be divided into $n_b$ mini-batches $\{ \mathcal{C}_{i_b} \}_{i_b=1}^{n_b}$, where every mini-batch $\mathcal{C}_{i_b}$ contains $m$ samples such that $N = m n_b$. Denoting the parameters at $i_b$-th iteration of a certain epoch $j$ as $\Theta_{i_b}^{(j)}$, for every mini-batch $\mathcal{C}_{i_b}$ and $\mb{x}^{(l)} \in \mathcal{C}_{i_b}, l = 1, \dots, m$, one can apply the mini-batch to estimate the expectation of the residual loss and the stochastic gradient, and then update the parameters $\Theta$ based on the following scheme
\begin{equation} \label{eq_sgd}
\Theta_{i_b}^{(j)} = \Theta_{i_b-1}^{(j)} -  \eta \nabla_{\Theta} \left[ \frac{1}{m} \sum\limits_{l=1}^m \left( r(\mb{x}^{(l)}; \Theta_{i_b-1}^{(j)}) \right)^2 \right] \quad \text{for} \ i_b = 1, \ldots, n_b, j = 1,2,\ldots
\end{equation} 
where $\eta$ is a given learning rate. Compared with the gradient descent method, the stochastic gradient descent method only requires computing the gradient on the mini-batch $\mathcal{C}_{i_b}$. \revs{In this work we employ the Adam optimizer, which is widely used to accelerate the training process for deep neural networks, as this method adopts adaptive learning rates for different components of parameters through estimates of first and second moments of the gradients \cite{kingma2014adam}}. 

\begin{figure}[ht] 
     \centering
     \subfloat[][FEM mesh]{\includegraphics[width=.45\textwidth]{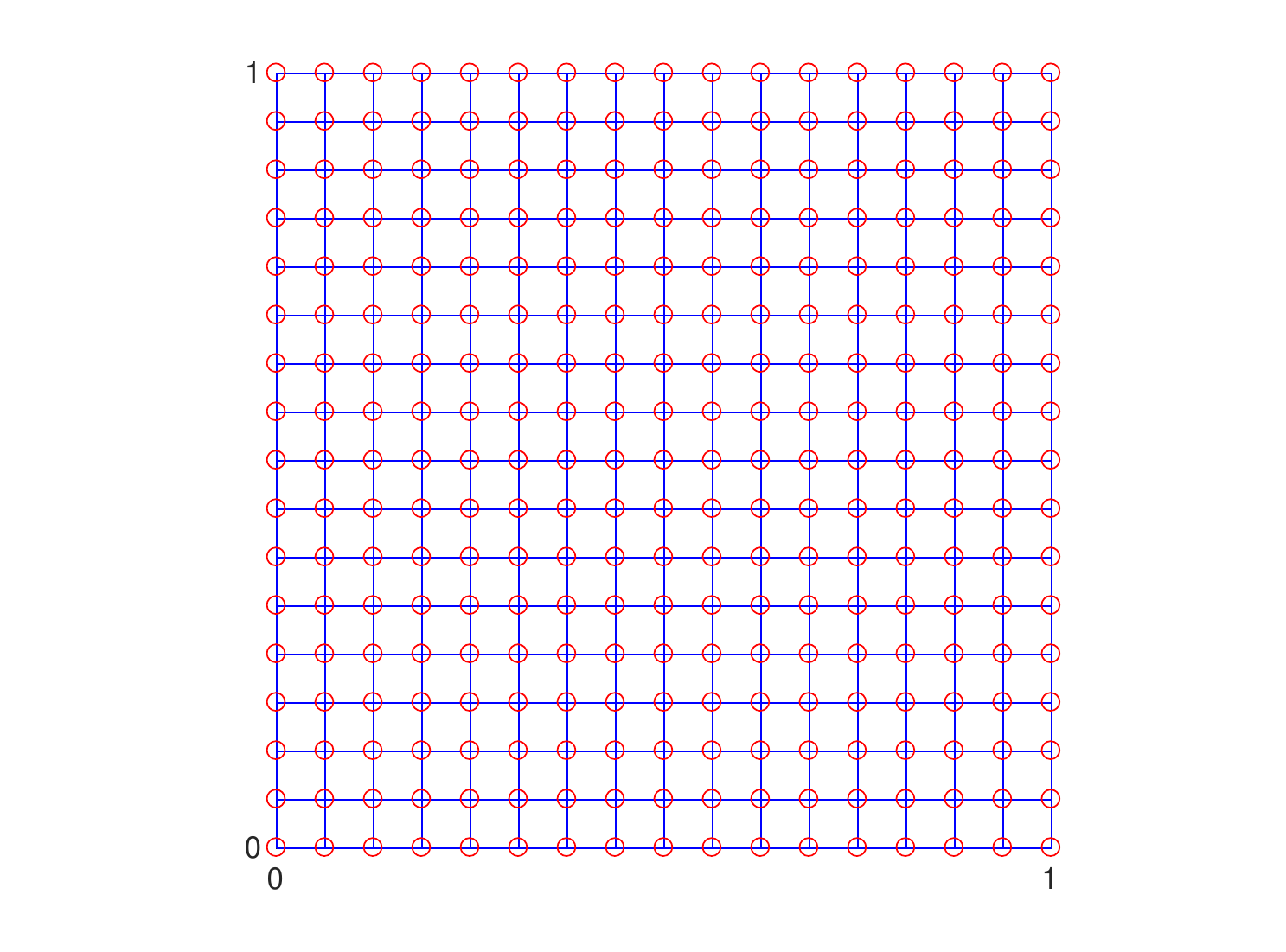}}\quad
     \subfloat[][random samples]{\includegraphics[width=.45\textwidth]{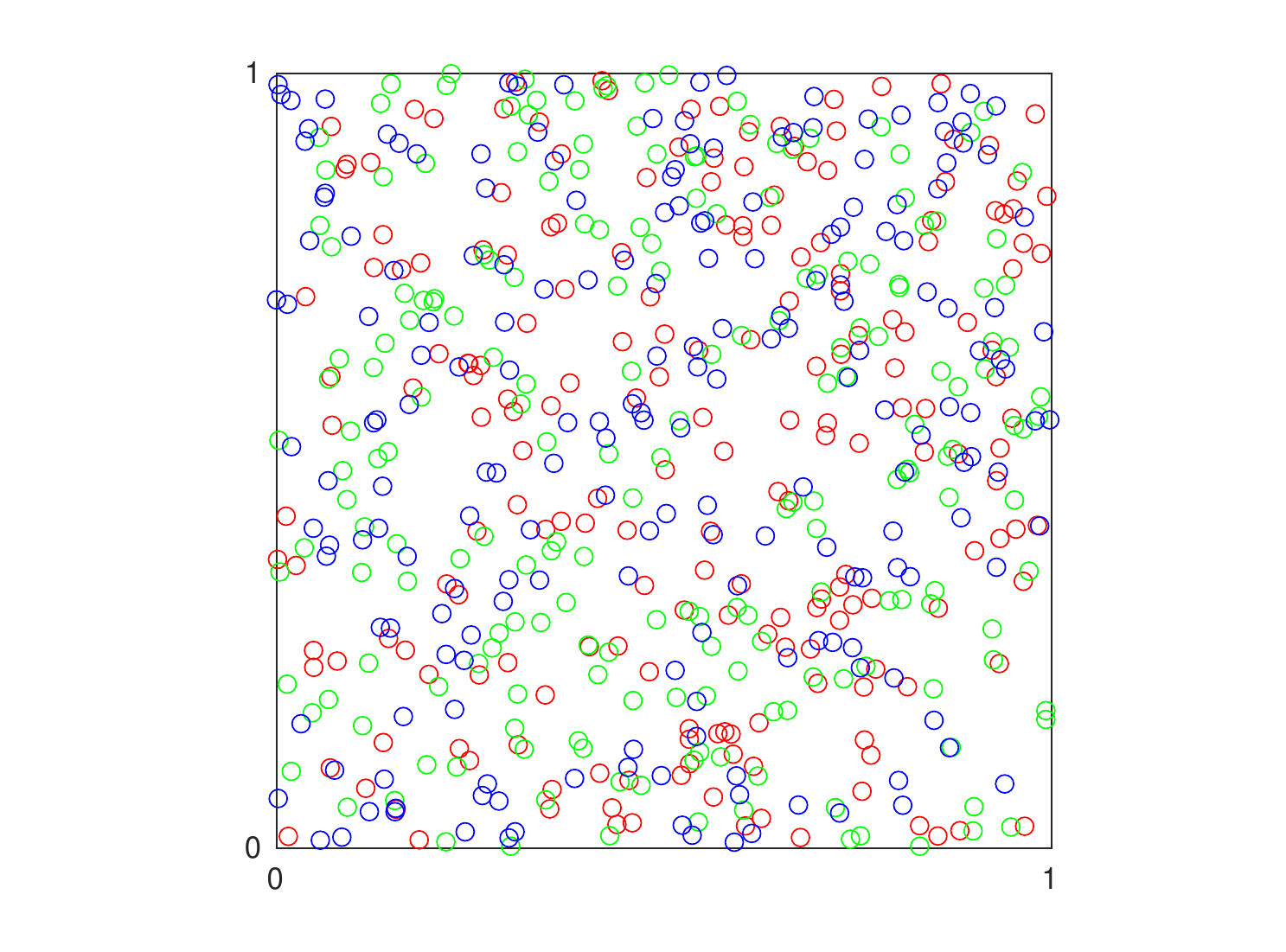}}
     \caption{An example of linear finite element meshes and stochastic collocation points in $[0, 1]^2$.}
     \label{fig_mesh}
  \end{figure}


\subsection{Adaptive sampling procedure} \label{sec_adap_fp}
Compared with the standard finite element methods (FEM) \cite{elman2014finite}, the deep learning approach does not require mesh generation to solve PDEs, which shares more similarities to meshless methods, and the approximation of Eq.\eqref{eq_resloss} fits naturally with \revs{stochastic gradient-based optimization}.  Figure \ref{fig_mesh} shows a linear finite element mesh in $[0,1]^2$ and the collocation points that are generated with a uniform distribution in $[0,1]^2$.

Adaptivity plays an important role in classical numerical methods for the approximation of PDEs.  Considering a finite element method subject to a certain mesh of the computation domain, we expect that the element-wise approximation errors are distributed in a nearly uniform way. This means that the most effective mesh should be non-uniform since the regularity of the solution varies in the computation domain. In our problem, the distribution $p_{\mathsf{data}}(\mb{x})$ of the collocation points will affect the approximation of $J(p_{X}(\mb{x};\Theta))$ and the optimal parameter $\Theta^*$ as well. Apparently a uniform distribution is not an optimal choice for $p_{\mathsf{data}}(\mb{x})$ especially for high-dimensional problems. For a certain amount of collocation points, the curse of dimensionality will weaken the contribution of each collocation point to our learning problem, which will be worsen for the approximation of PDF if the exact solution $p(\mb{x})$ is far away from being uniform. We then expect to use samples from a nonuniform distribution $p_{\mathsf{data}}(\mb{x})$ for the approximation $J(p_{X}(\mb{x};\Theta))$, where a simple criterion is that  $p_{\mathsf{data}}(\mb{x})$ should be consistent with the true solution $p(\mb{x})$ to some extent. This will result in adaptive deep density approximation (ADDA) for the approximation of the Fokker-Planck equation.


It is, in general, difficult to generate samples that are adaptive to the true solution $p(\mb{x})$. Fortunately, flow-based deep generative models provide an opportunity for us to do this thanks to the invertible mapping. 
Our strategy is as follows. 
Starting with an initial set of collocation points $\mathcal{C}_0 =\left \{ \mb{x}_{(0)}^{(i)}\right\}_{i=1}^{N}$ drawn from a uniform distribution, 
we train and obtain the KRnet $Z=f_{\mathsf{KR},(0)}(X;\Theta^{*,(0)})$, which corresponds to the PDF $p_X^{(0)}(\mb{x};\Theta^{*,(0)})$. 
We then generate a new set $\mathcal{C}_1 = \left\{ \mb{x}_{(1)}^{(i)}\right\}_{i=1}^{N}$ of collocation points by $X=f^{-1}_{\mathsf{KR},(0)}(Z)$ using $N$ samples from the prior distribution of $Z$. Then $C_1$ is a set of samples from $p_X^{(0)}(\mb{x};\Theta^{*,(0)})$. We continue to update the KRnet using  $\Theta^{*,(0)}$ as the initial parameters and $C_1$ as the training set, which yields $f_{\mathsf{KR},(1)}(X;\Theta^{*,(1)})$. Then another iteration starts. In general, we sample the current optimal PDF model $p_{X}^{(k)}(\mb{x};\Theta^{*,(k)})$ to generate a new training set $C_{k+1}=\left\{ \mb{x}_{(k+1)}^{(i)}\right\}_{i=1}^{N}$ and update the KRnet to $f_{\mathsf{KR,(k+1)}}(\mb{x};\Theta^{*,(k+1)})$. 
This way, the samples for the training process become more and more consistent with the true solution, if $p_X^{(k)}(\mb{x};\Theta^{*,(k)})$ approaches $p(\mb{x})$ as $k$ increases. In other words, more collocation points will be chosen in the region of high density while less collocation points in the region of low density. Our adaptive training process has been summarized in Algorithm \ref{alg_2}, where $N_{\rm adaptive}\in \xs{N}$ is  a given number of  maximum adaptivity iterations,
	and this strategy is called the adaptive deep density approximation based on KRnet (ADDA-KR) from now on.
	The final KRnet-induced PDF  
	is the ADDA-KR approximation for the  steady state Fokker-Planck problem 
	\eqref{eqn_fpsta}--\eqref{eq_pdf_positivity}. 

We note that the adaptivity in Algorithm \ref{alg_2} can be further tuned. One possible strategy is to update the training set gradually for each training stage, e.g., up to a certain percentage. In this work, we replace the whole training set from the previous stage just for simplicity.

\subsection{Implementation issues}
When minimizing the loss functional Eq.\eqref{eqn_fploss}, numerical underflow issues can be encountered, 
especially when $\mb{x}$ is relatively high-dimensional. That is, 
the loss functional can be too small to provide an effective gradient descent direction.
To alleviate this issue, we develop the following scaling strategy in our implementation. 
Multiplying both sides of equation \eqref{eqn_fpsta} by a constant $C_s > 0$ gives
\begin{equation} \label{eqn_fpsta_scale}
\mathcal{L} \left( C_sp(\mb{x}) \right)= \nabla \cdot \left [C_sp(\mb{x}) \nabla V(\mb{x}) \right ] + \nabla \cdot [\nabla \cdot (C_sp(\mb{x}) \mb{D}(\mb{x}))] = 0.
\end{equation}	
The solution of the above equation is the same as the solution of the original stationary Fokker-Planck equation \eqref{eqn_fpsta}.
However, if $C_s$ is large enough, Eq.\eqref{eqn_fpsta_scale} is numerically more stable than  Eq.\eqref{eqn_fpsta},
and the loss functional Eq.\eqref{eqn_fploss} associated with Eq.\eqref{eqn_fpsta_scale} can typically provide effective gradient descent directions 
to optimize the parameters $\Theta$. 
In our practical implementation, we usually set $C_s = 100$.

%

\begin{algorithm}
	\caption{Adaptive deep density approximation based on KRnet (ADDA-KR)  for the Fokker-Planck equation}
	\label{alg_2}
	\begin{algorithmic}[1]
		\Require Initial KRnet $p_X^{(0)}(\mb{x};\Theta_{0}^{(0)})$, 
		maximum epoch number $N_e$, maximum iteration number $N_{\rm adaptive}$, learning rate $\eta$, batch size $m$, and initial training set $\mathcal{C}_0 = \left\{ \mb{x}_{(0)}^{(i)} \right\}_{i=1}^{N}$.
		\State Divide $\mathcal{C}_0 = \left\{ \mb{x}_{(0)}^{(i)}  \right\}_{i=1}^{N}$ into $n_b$ mini-batch $\left\{ \mathcal{C}_{i_b} \right\}_{i_b=1}^{n_b}$.
		\For {$k=1:N_{\rm adaptive}$}
		\For {$j = 0:N_e-1$}
		\For {$i_b = 1:n_b$}
		\State Compute the values of the residual loss 
		$ r(\mb{x}_{(k-1)}^{(l)}; \Theta_{i_b -1}^{(j)})$ (see Eq.\eqref{eqn_fploss}) for $l=1,\ldots,m$, on the mini-batch $\mathcal{C}_{i_b}$.
		\State Update the parameters \revs{$\Theta_{i_b}^{(j)}$ using the Adam optimizer with learning rate $\eta$.}
		\EndFor
		\If {$j = N_e - 1$}
		\State Let $\Theta^{*, (k)} := \Theta_{n_b}^{(N_e - 1)}$.
		\Else
		\State Let $\Theta_{0}^{(j+1)} := \Theta_{n_b}^{(j)}$.
		\EndIf
		\State Shuffle the set of collocation points $\mathcal{C}_{k-1} = \left\{ \mb{x}_{(k-1)}^{(i)}  \right\}_{i=1}^{N}$.
		\State Divide $\mathcal{C}_{k-1} = \left\{ \mb{x}_{(k-1)}^{(i)}  \right\}_{i=1}^{N}$ into $n_b$ mini-batch $\left\{ \mathcal{C}_{i_b} \right\}_{i_b=1}^{n_b}$.
		\EndFor
		\If {$k = N_{\rm adaptive}$} 
		\State Let $\Theta := \Theta^{*, (k)} $.
		\Else
		\State Generate $\mathcal{C}_{k+1} = \left\{ \mb{x}_{(k+1)}^{(i)}  \right\}_{i=1}^{N}$ by $p_X^{(k)}(\mb{x};\Theta^{*, (k)})$.
		\State Let $\Theta_{0}^{(0)} := \Theta^{*, (k)}$.
		\EndIf
		\EndFor
		\State Obtain the  ADDA-KR solution $p_X(\mb{x};\Theta) := p_X^{(N_{\rm {adaptive}})}(\mb{x};\Theta) $.
		
		\Ensure The ADDA-KR solution $p_X(\mb{x};\Theta)$.
	\end{algorithmic}
\end{algorithm}

%% file: section6.tex
In this section, numerical experiments are conducted to illustrate the effectiveness of our ADDA-KR (adaptive deep density approximation based on KRnet) approach presented in Algorithm \ref{alg_2}. 
 Five test problems for the Fokker-Planck equation are studied---one one-dimensional test problem, 
 two two-dimensional test problems (one is a single modal distribution, and the other is a bimodal distribution),
 one four-dimensional test problem, and one eight-dimensional test problem. 
The activation function of $\mathsf{NN}_{[i]}$ (see Eq.\eqref{eq_nn_affine}) is set to the hyperbolic tangent function for all test problems.
For comparison, we also test the performance of a direct adaptive version of classic real NVP, 
and as the real NVP utilizes a half-half partition (see section \ref{sec_KRnet_density}), we refer to it as ADDA-HH. 
The implementation of ADDA-HH is to replace the KRnet in  ADDA-KR (Algorithm \ref{alg_2}) by 
the classical real NVP, and we set the same input parameters for both ADDA-KR and ADDA-HH in all our test problems. 
In addition, results of non-adaptive versions of KRnet and real NVP are included for 
high-dimensional test problems (the four-dimensional and the eight-dimensional test problems),
which are referred to as Uniform-KR and Uniform-HH. In Uniform-KR and Uniform-HH, collocation points are 
generated through uniform distributions, and other settings of KRnet and real NVP are the same as the settings for 
ADDA-KR in these test problems. 

\subsection{A one-dimensional test problem} \label{sec_test_1dou}
We start with this one-dimensional case, 
where the governing equation is 
\begin{equation}
\begin{aligned}
\frac{\partial (x p(x))}{\partial x} + \frac{1}{2} \frac{\partial^2 ( p(x))}{\partial x^2} &= 0,\\
\int_{\xs{R} } p(x) dx = 1, \ p(x) &\geq 0,  \\
\end{aligned}
\end{equation}
and the exact 
solution is 
\begin{equation}
p(x) = \frac{\mathrm{exp}(-x^2)}{\sqrt{\pi}}.
\end{equation}
For this one-dimensional problem, KRnet is the same as the classical real NVP, 
meaning that only the affine coupling layers are needed.  
As the affine coupling layers (see section \ref{sec_coupling_layer}) need at least two-dimensions,
we use $[x, x]$ as an input in our implementation of KRnet.  
We generate  the initial parameters $\Theta_{0}^{(0)}$ for the inputs of Algorithm \ref{alg_2},
using Glorot Gaussian initialization \cite{glorot2010understanding},
and  then construct the initial KRnet $p_X^{(0)}(\mb{x};\Theta_{0}^{(0)})$. 
The number of epochs is set to $N_e=300$, 
and only one adaptivity iteration is conducted for this one-dimensional problem,
i.e., $N_{\rm adaptive}=1$. 
 The learning rate for Adam optimizer is set to $\eta=0.0002$, 
 and the batch size is set to $m = 500$. 
The initial training set $\mathcal{C}_0$ is generated through the uniform distribution with range $[-5,5]$, 
 and the sample size is set to  $|\mathcal{C}_k|=3000$ for each iteration step $k$ for $k=0,\ldots,N_{\rm adaptive}$.
In addition, we take $L = 8$ affine coupling layers, and two fully connected layers with $w = 48$ neurons for $\mathsf{NN}_{[i]}$ (see Eq.\eqref{eq_nn_affine}). 



\begin{figure}
	\center{
		\includegraphics[width=0.6\textwidth]{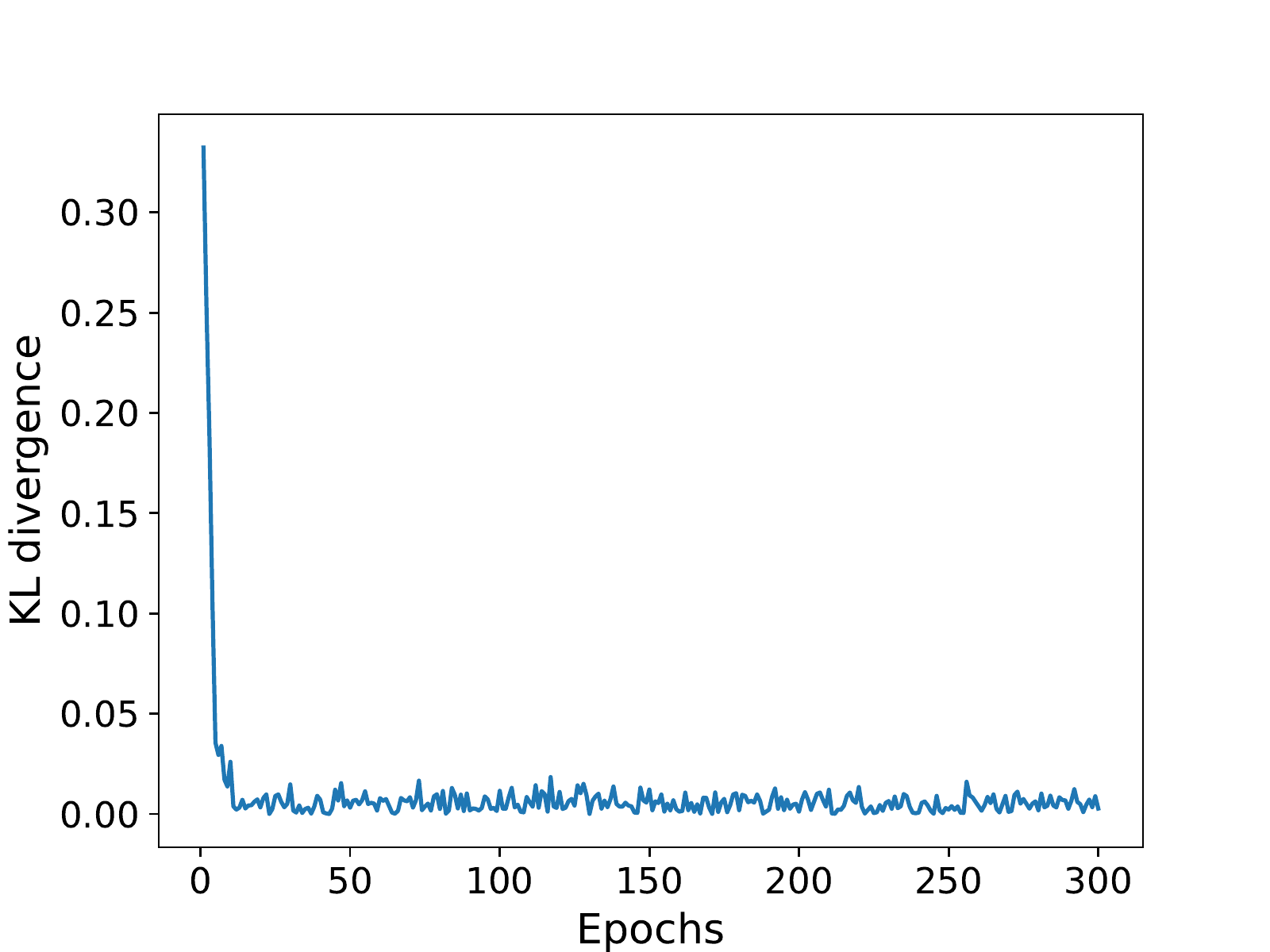}
	}
	\caption{The KL divergence with respect to epochs, one-dimensional test problem.}\label{fig:ou1d_kld}
\end{figure}
\begin{figure}
	\center{
		\includegraphics[width=0.6\textwidth]{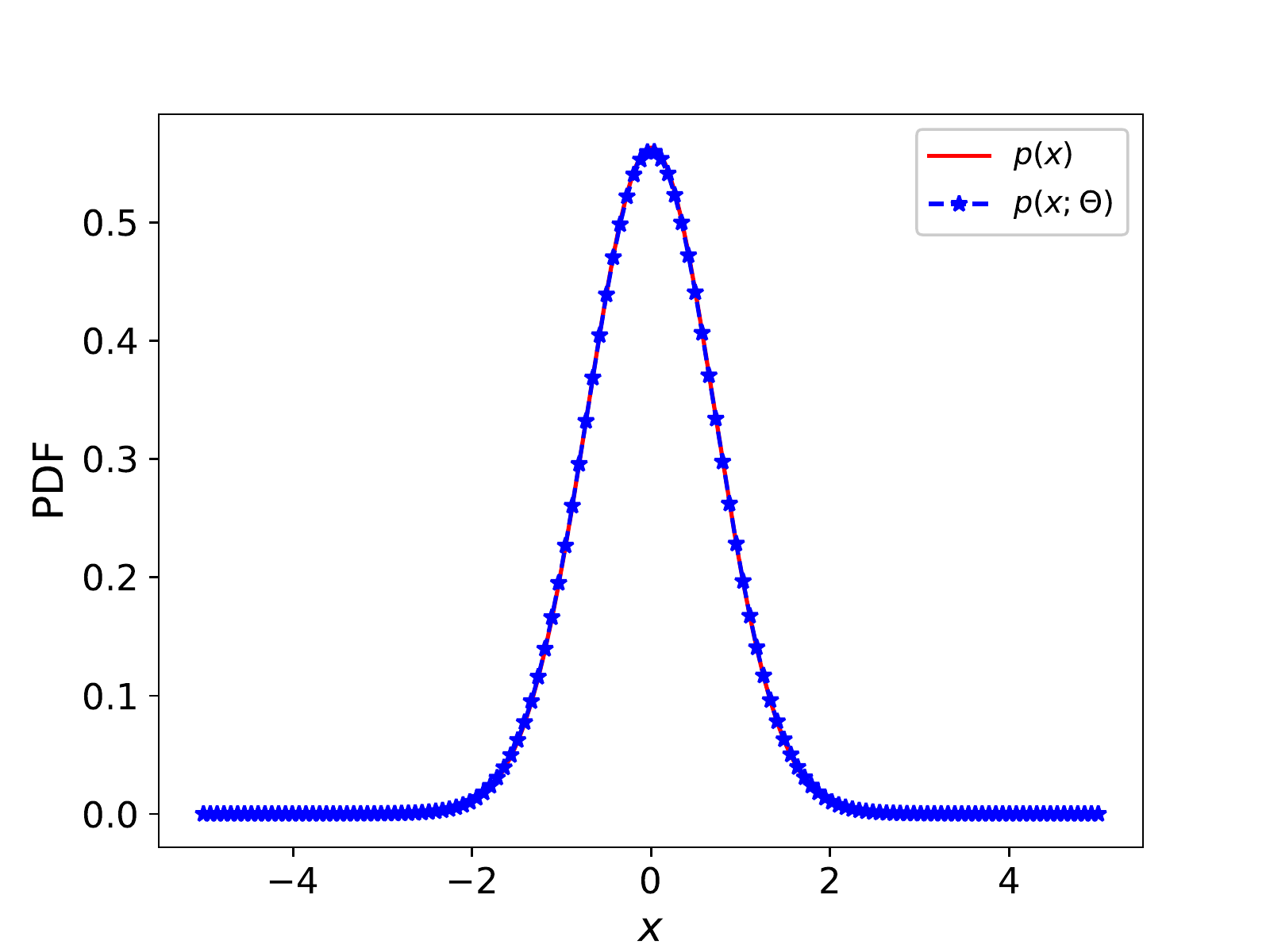}
	}
	\caption{The exact solution and the ADDA-KR solution, one-dimensional test problem.}\label{fig:ou1d_pdf}
\end{figure}

To assess the accuracy of our ADDA-KR approach (Algorithm \ref{alg_2}), we compute the KL divergence between 
the exact solution $p(x)$ 
and our ADDA-KR solution $p_X(x;\Theta)$: 
\begin{equation*}
\begin{aligned}
D_{KL}(p(x)|| p_X(x;\Theta)) &= \int_{-\infty}^{\infty} p(x) \log p(x) dx - \int_{-\infty}^{\infty} p(x) \log p_X(x;\Theta) dx \\
& = -\frac{1}{2}(1 + \log \pi) - \int_{-\infty}^{\infty} p(x) \log p_X(x;\Theta) dx
\end{aligned}
\end{equation*}
where the last term of the above equation is approximated by Monte Carlo integration with $10^4$ samples.
Figure \ref{fig:ou1d_kld} shows the KL divergence decreases  to zero quickly. 
Figure \ref{fig:ou1d_pdf} shows the  exact solution $p(x)$  and our ADDA-KR solution $p_X(x;\Theta)$,
where it can be seen that they are visually  indistinguishable. 
	
\subsection{Two-dimensional test problems} \label{sec_test_2dfp}
In this part, two-dimensional Fokker-Planck equations are considered, where the solution of the first one
is a single modal distribution and the solution of the second one is a bimodal distribution. 

\subsubsection{Two-dimensional single modal distribution}
The stationary Fokker-Planck equation for this test problem is
\begin{equation} \label{eq_oufp}
\begin{aligned}
\nabla \cdot \left [p(\mb{x}) \mb{A} \mb{x} \right ] + &\nabla \cdot [\nabla \cdot (p(\mb{x}) \mb{D}] = 0, \\
\int_{\xs{R}^d } p(\mb{x}) d \mb{x} &= 1, \ p(\mb{x}) \geq 0, \\
\end{aligned}
\end{equation} 
where $\mb{A}$ and $\mb{D}$ are two constant matrices. This equation is corresponding to the following Ornstein-Uhlenbeck process
\begin{equation}
d X_t = -\mb{A} X_t dt + \mb{G} d \mathbf{w}_t,
\end{equation}
where  $\mb{D} = \mb{G} \mb{G}^\mathsf{T}/2$. 

 The solution of Eq.\eqref{eq_oufp} exists if the real parts of the eigenvalues of $\mb{A}$ are larger than zero \cite{risken1984fokker},
 and it can be written as 
\begin{equation}
p(\mb{x}) = (2 \pi)^{-1} (\mathrm{det} \mb{\Sigma})^{-\frac{1}{2}} \mathrm{exp}(-\frac{1}{2}\bx^\mathsf{T} \mathbf{\Sigma}^{-1} \bx),
\end{equation}
where the covariance matrix $\mathbf{\Sigma}$ is determined by the following Lyapunov equation
\begin{equation}
\mb{A} \mb{\Sigma} + \mb{\Sigma} \mb{A }^\mathsf{T} = 2 \mb{D}.
\end{equation}
The above Lyapunov equation has a unique solution if and only if the eigenvalues $\lambda_i$ of $\mb{A}$ satisfy $\lambda_i \neq -\lambda_j$ for all $i, j = 1, 2$. In this test problem, the constant matrix $\mb{A}$ for the drift term and the diffusion matrix $\mb{D}$ are
set to   
\begin{equation*}
\mb{A} = \left[ \begin{array}{cc}
         1.37096037 & -0.48306187  \\
         -0.48306187 & 1.62903963
    \end{array}
    \right], \quad
\mb{D} = \left[ \begin{array}{cc}
         22.52429192 & -6.55821381  \\
         -6.55821381 & 12.68972
    \end{array}
    \right], 
\end{equation*}
which implies that the covariance matrix $\mb{\Sigma}$ is 
\begin{equation*}
\mb{\Sigma} = \left[ \begin{array}{cc}
         8.12186142 & -0.26372569  \\
         -0.26372569 & 3.81664391
\end{array}
\right].
\end{equation*}

We generate  the initial parameters $\Theta_{0}^{(0)}$ with Glorot Gaussian initialization \cite{glorot2010understanding},
and  then construct the initial KRnet $p_X^{(0)}(\mb{x};\Theta_{0}^{(0)})$ for Algorithm \ref{alg_2}. 
The number of epochs is set to $N_e=300$, 
and two adaptivity iterations are conducted for this problem,
i.e., $N_{\rm adaptive}=2$. 
The learning rate for Adam optimizer is set to $\eta=0.0002$, 
and the batch size is set to $m = 1000$. 
The initial training set $\mathcal{C}_0$ is generated through the uniform distribution with range $[-6,6]^2$, 
and the sample size is set to  $|\mathcal{C}_k|=6 \times 10^4$ for each iteration step $k$ for $k=0,\ldots,N_{\rm adaptive}$.
In addition, we take $L = 8$ affine coupling layers, and two fully connected layers with $w = 48$ neurons for $\mathsf{NN}_{[i]}$ (see Eq.\eqref{eq_nn_affine}). 

\begin{figure}[ht]
     \centering
     \subfloat[][Exact solution $p(\mb{x})$]{\includegraphics[width=.48\textwidth]{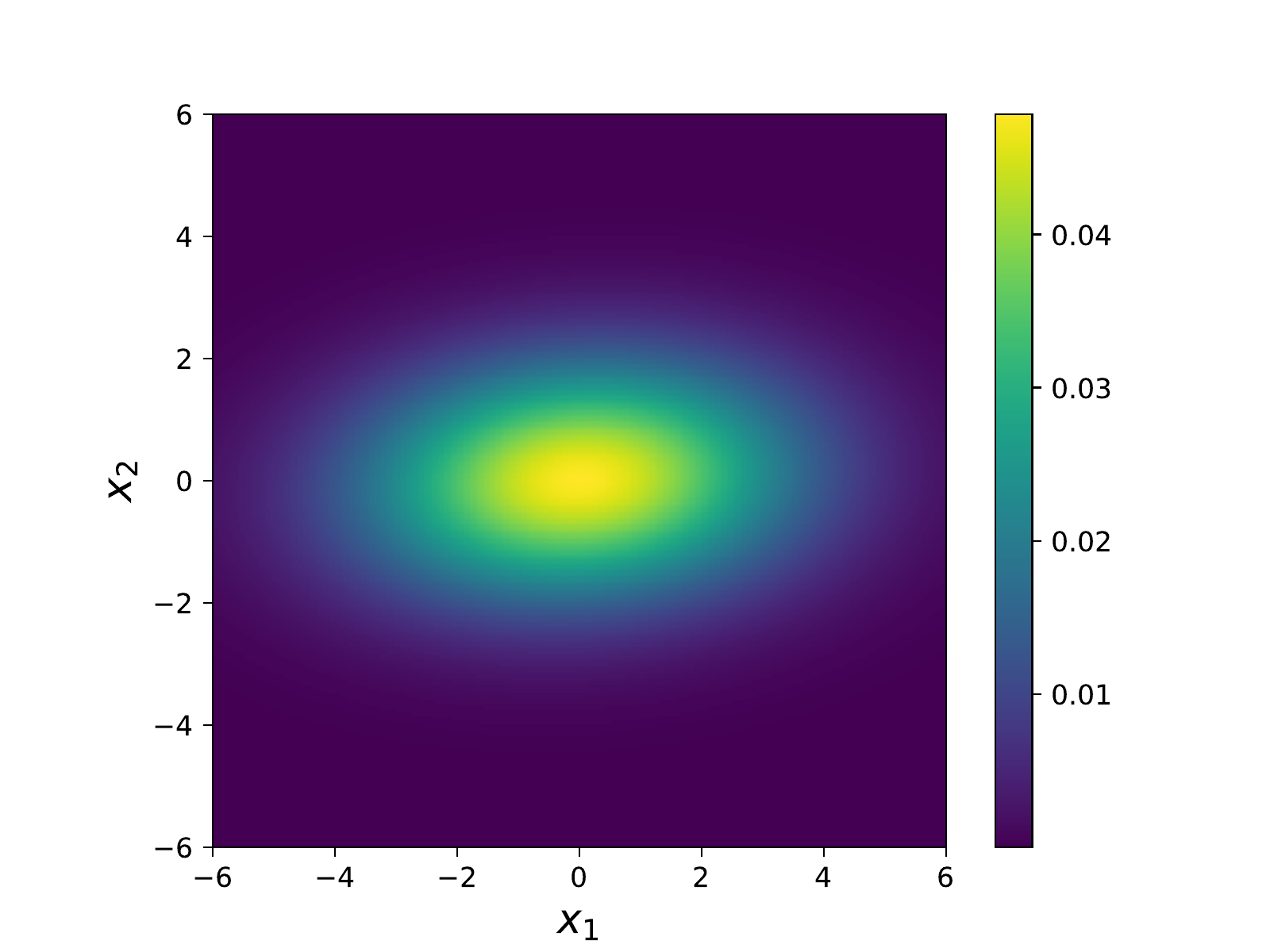}}\quad
     \subfloat[][ADDA-KR solution $p_{X}(\mb{x};\Theta)$]{\includegraphics[width=.48\textwidth]{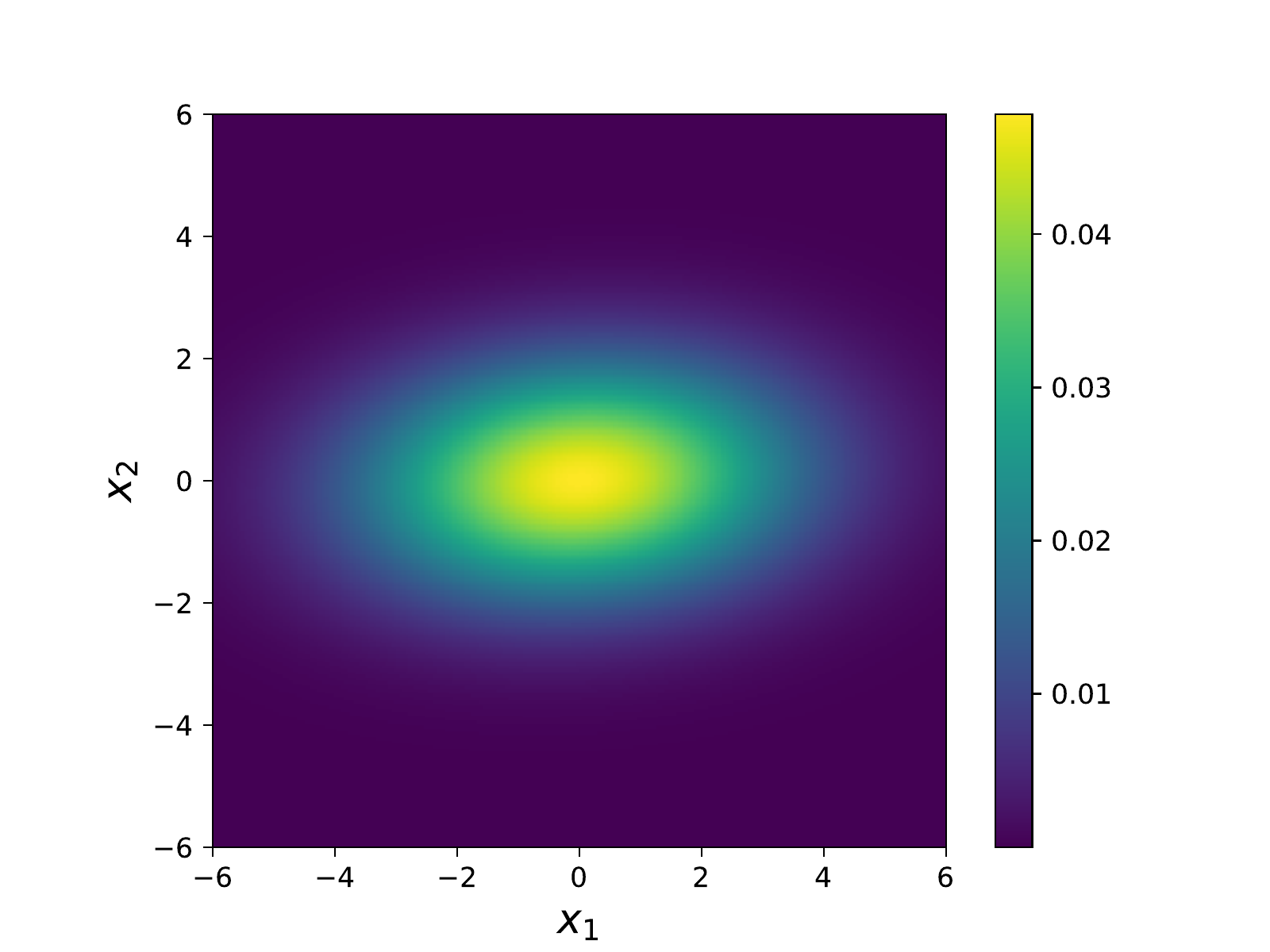}}
     \caption{Solutions, two-dimensional single modal test problem.}
     \label{fig:ou2d_pdf}
  \end{figure}
\begin{figure}[ht]
     \centering
     \subfloat[][Samples of the exact solution $p(\mb{x})$.]{\includegraphics[width=.45\textwidth]{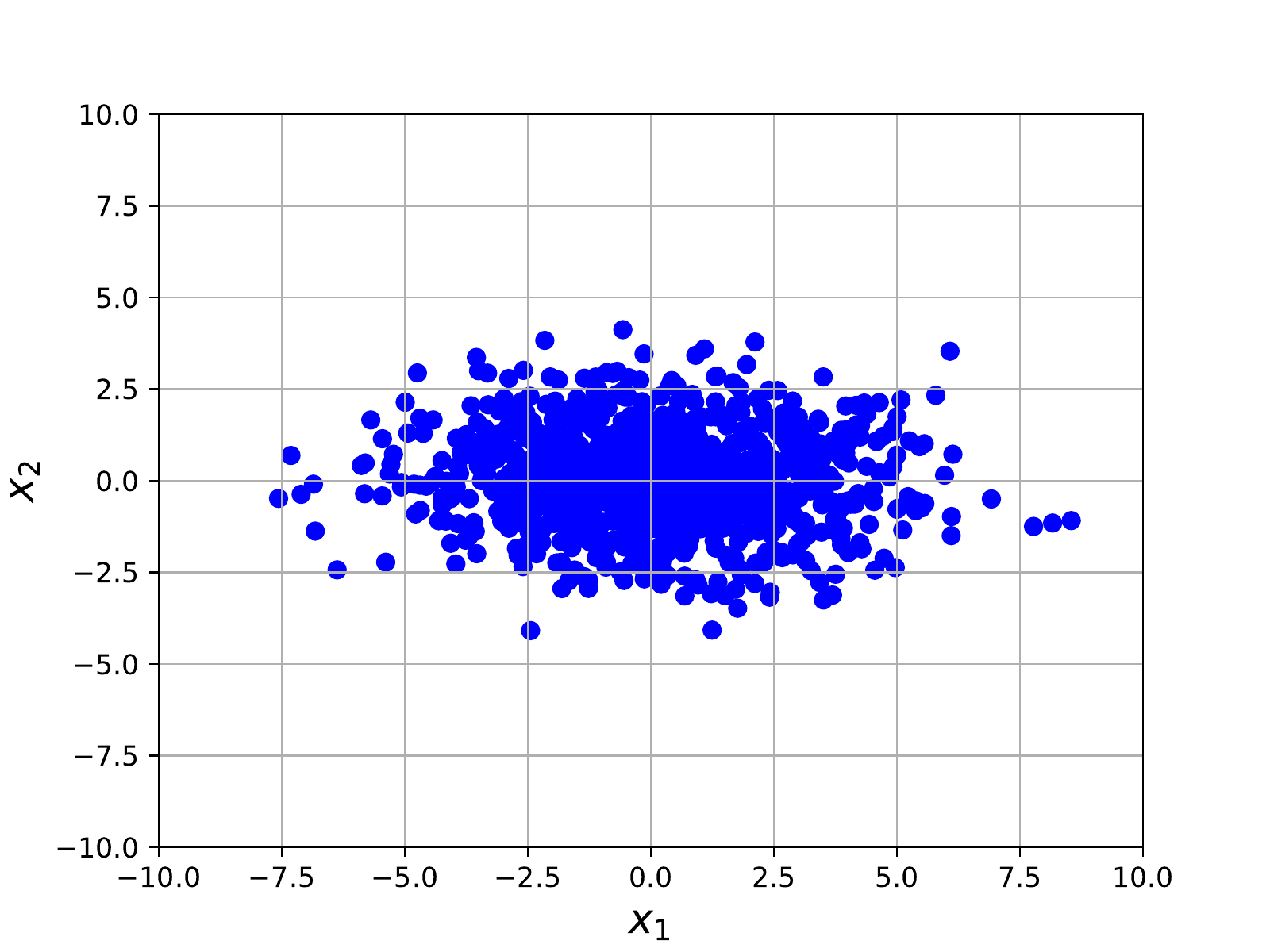}}\quad
     \subfloat[][Samples of the ADDA-KR solution $p_{X}(\mb{x};\Theta)$.]{\includegraphics[width=.45\textwidth]{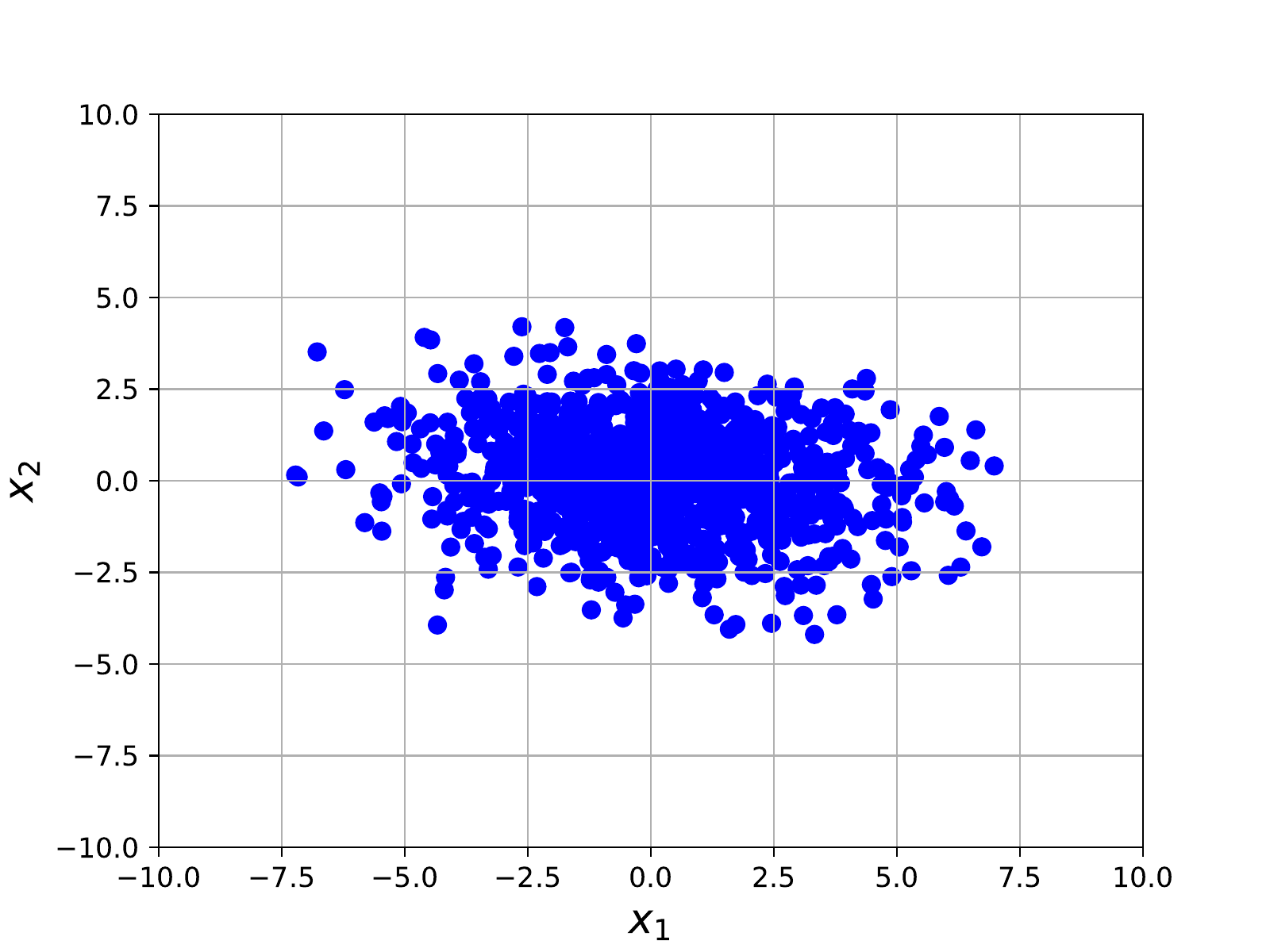}}
     \caption{Samples, two-dimensional single modal test problem.}
     \label{fig:ou2d_sample}
  \end{figure}

Figure \ref{fig:ou2d_pdf} shows the exact solution $p(\mb{x})$ and our ADDA-KR solution $p_X(\mb{x};\Theta)$, where it can be seen that they are visually indistinguishable. 
For this test problem, there is no significant difference between the ADDA-KR solution  and the ADDA-HH solution,
and we then only show the  exact solution and our ADDA-KR solution. 
Figure \ref{fig:ou2d_sample} shows samples drawn from the exact  solution of  Eq.\eqref{eq_oufp} and our ADDA-KR solution, 
which confirms that the corresponding distributions ($p(x)$ and $p_X(\mb{x};\Theta)$) are very close.



\subsubsection{Two-dimensional bimodal distribution}
In this test problem, the Fokker-Planck equation considered is 
\begin{equation} \label{eq_testfp_bimodal}
\begin{aligned}
 -\nabla \cdot \left [p(\mb{x}) \nabla \log (\beta_1 p_1(\mb{x}) + \beta_2 p_2(\mb{x})) \right ] &+ \nabla^2 p(\mb{x}) = 0, \\
\int_{\xs{R}^d } p(\mb{x}) d \mb{x} &= 1, \ p(\mb{x}) \geq 0,\\
\end{aligned}
\end{equation} 
where for $k=1,2$, each $p_k(\mb{x})$ is the probability density function of the normal distribution with mean $\mathbf{\mu}_k$ and covariance $\mathbf{\Sigma}_k$, 
and $\beta_1 + \beta_2 = 1$. 
The solution of Eq.\eqref{eq_testfp_bimodal} is the following Gaussian mixture distribution  \cite[p.~123]{pavliotis2014stochastic}, 
\begin{equation}\label{eq_bisol}
p(\mb{x})= \beta_1 p_1(\mb{x}) + \beta_2 p_2(\mb{x}).
\end{equation}
Here, we set $\mb{\mu}_k$, $\mb{\Sigma}_k$ and $\beta_k$ for $k=1,2$ as 
\begin{equation} \label{eq_2dmix_setting}
\begin{aligned}
\beta_1 = 0.55, \ \beta_2 = 0.45, &\quad \mb{\mu}_1 = [-1,-1]^\mathsf{T}, \  \mb{\mu}_2 = [2, 2]^\mathsf{T} \\
\mb{\Sigma}_1 = \left[ \begin{array}{cc}
         6.12186142 & -0.26372569  \\
         -0.26372569 & 1.81664391
\end{array}
\right],  &\quad \mb{\Sigma}_2 = \left[ \begin{array}{cc}
         2.8828528 & -0.70234742  \\
         -0.70234742 & 2.69199911
\end{array}
\right].
\end{aligned}
\end{equation} 
The matrices $\mb{\Sigma}_1$ and $\mb{\Sigma}_2$ are positive definite,
and their entries are randomly constructed.

We again generate  the initial parameters $\Theta_{0}^{(0)}$ with Glorot Gaussian initialization,
and  then construct the initial KRnet $p_X^{(0)}(\mb{x};\Theta_{0}^{(0)})$. 
The number of epochs is set to $N_e=200$, 
and the maximum number of adaptivity iterations conducted for this problem is set to $N_{\rm adaptive}=5$. 
The learning rate for Adam optimizer is set to $\eta=0.0001$, 
and the batch size is set to $m = 1000$. 
The initial training set $\mathcal{C}_0$ is generated through the uniform distribution with range $[-5,5]^2$, 
and the sample size is set to  $|\mathcal{C}_k|=6 \times 10^4$ for each iteration step $k$ for $k=0,\ldots,N_{\rm adaptive}$.
In addition, 
we take $L = 8$ affine coupling layers for both KRnet and real NVP, and two fully connected layers with $w = 48$ neurons for $\mathsf{NN}_{[i]}$ (see Eq.\eqref{eq_nn_affine}). For KRnet, we set $K=2$ to focus on the effectiveness of the rotation layer and 
the nonlinear layer for this test problem. 
To assess the effectiveness of our ADDA-KR approach, 
we generate a validation data set $\mathcal{C}_v = \{\mb{x}^{(i)}\}_{i=1}^{N_v}$, and compute the relative error defined by Eq.\eqref{eq_relative_error}. The KL divergence is approximated by Monte Carlo integration
\begin{equation} \label{eq_kl_error}
D_{KL}(p(\mb{x})||p_X(\mb{x};\Theta)) \approx \frac{1}{N_v}\sum\limits_{i=1}^{N_v} \left( \log p(\mb{x}^{(i)}) - \log p(\mb{x}^{(i)};\Theta) \right),
\end{equation}
where $\mb{x}^{(i)}$ is drawn from the exact solution $p(\mb{x})$, 
and the size of the validation data set is set to $3.2 \times 10^5$ such that the KL-divergence can be approximated well. 



%

\begin{figure}
	\center{
		\includegraphics[width=0.6\textwidth]{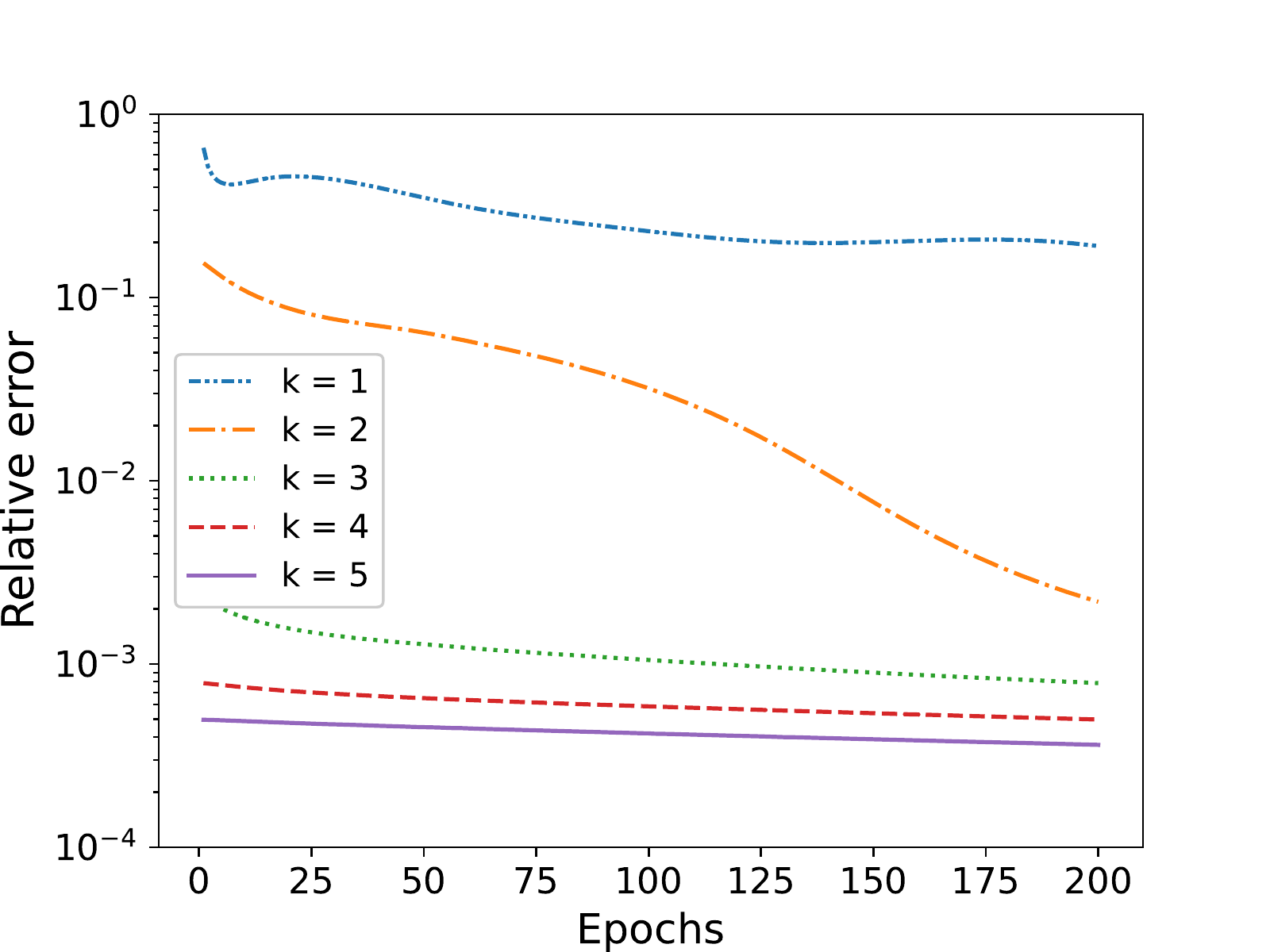}
	}
        \caption{The relative error for ADDA-KR,  two-dimensional bimodal test problem.}\label{fig:mix2d_k_convergence}
\end{figure}

\begin{figure}[!ht]
	\centering
	\subfloat[][Relative error w.r.t. adaptivity iteration step $k$. ]{\includegraphics[width=.4\textwidth]{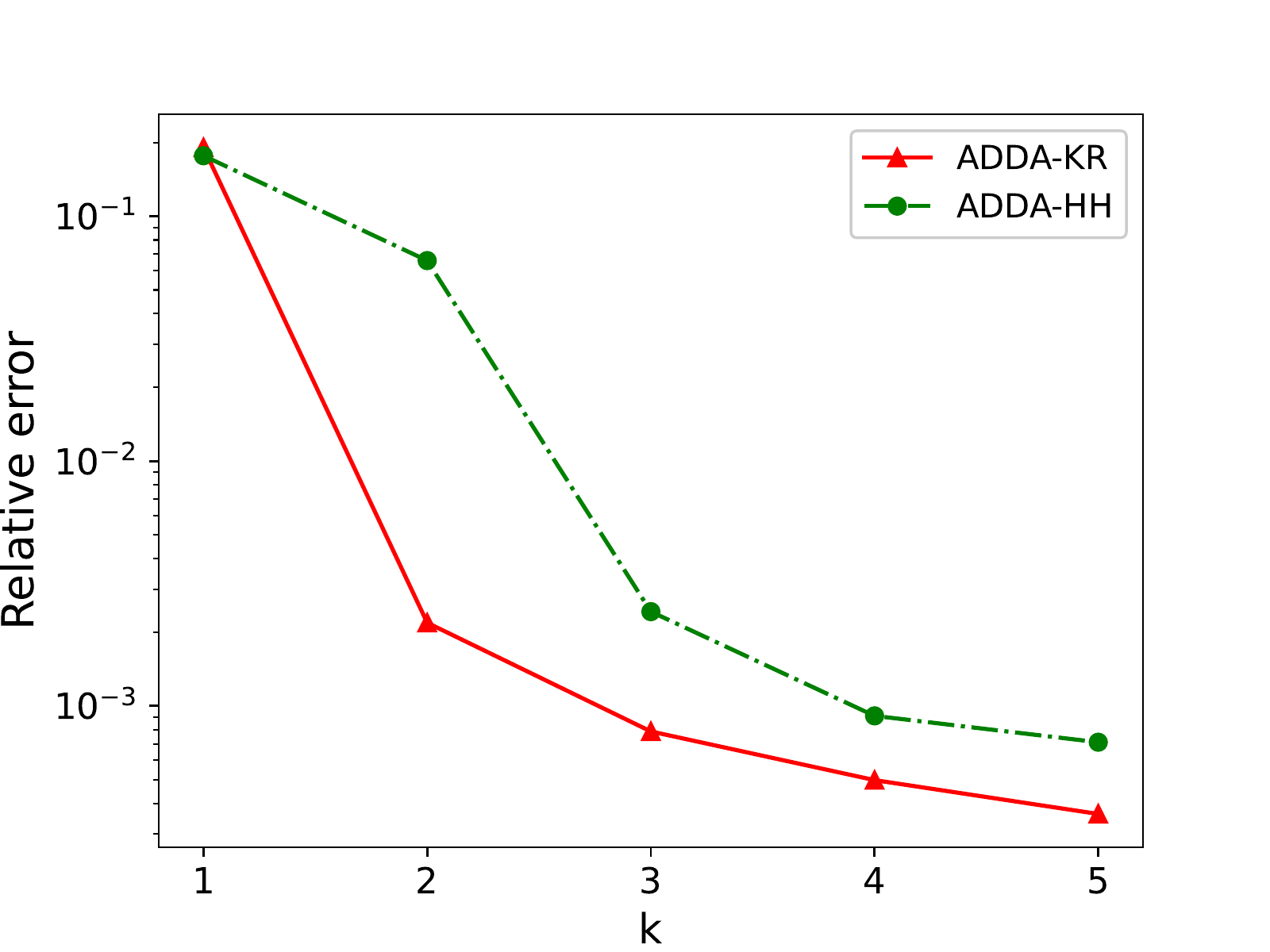}}\quad
	\subfloat[][Relative error at $k = 1$. ]{\includegraphics[width=.4\textwidth]{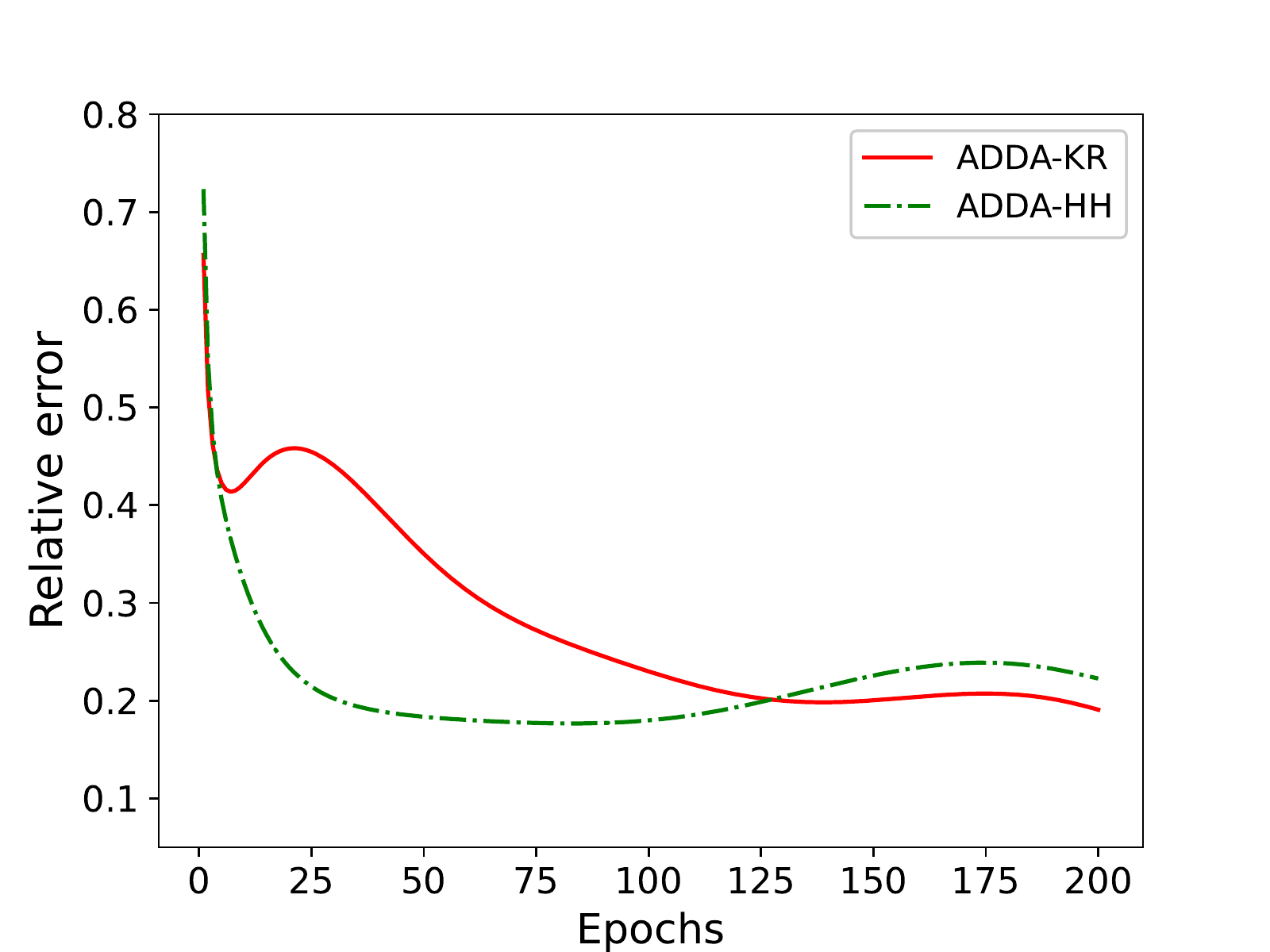}}\\
	\subfloat[][Relative error at $k = 3$. ]{\includegraphics[width=.4\textwidth]{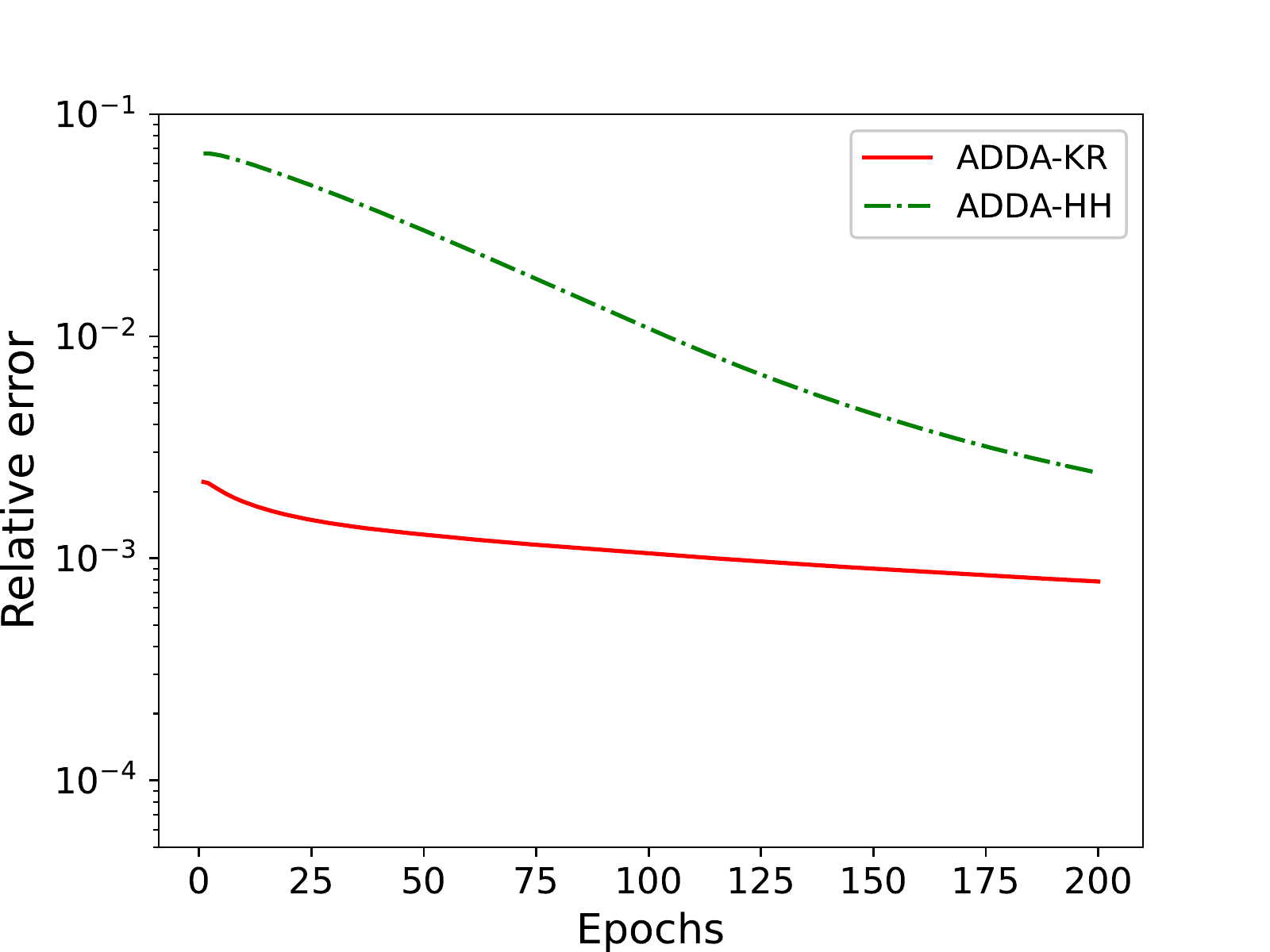}}\quad
	\subfloat[][Relative error at $k = 5$. ]{\includegraphics[width=.4\textwidth, height=.27\textwidth]{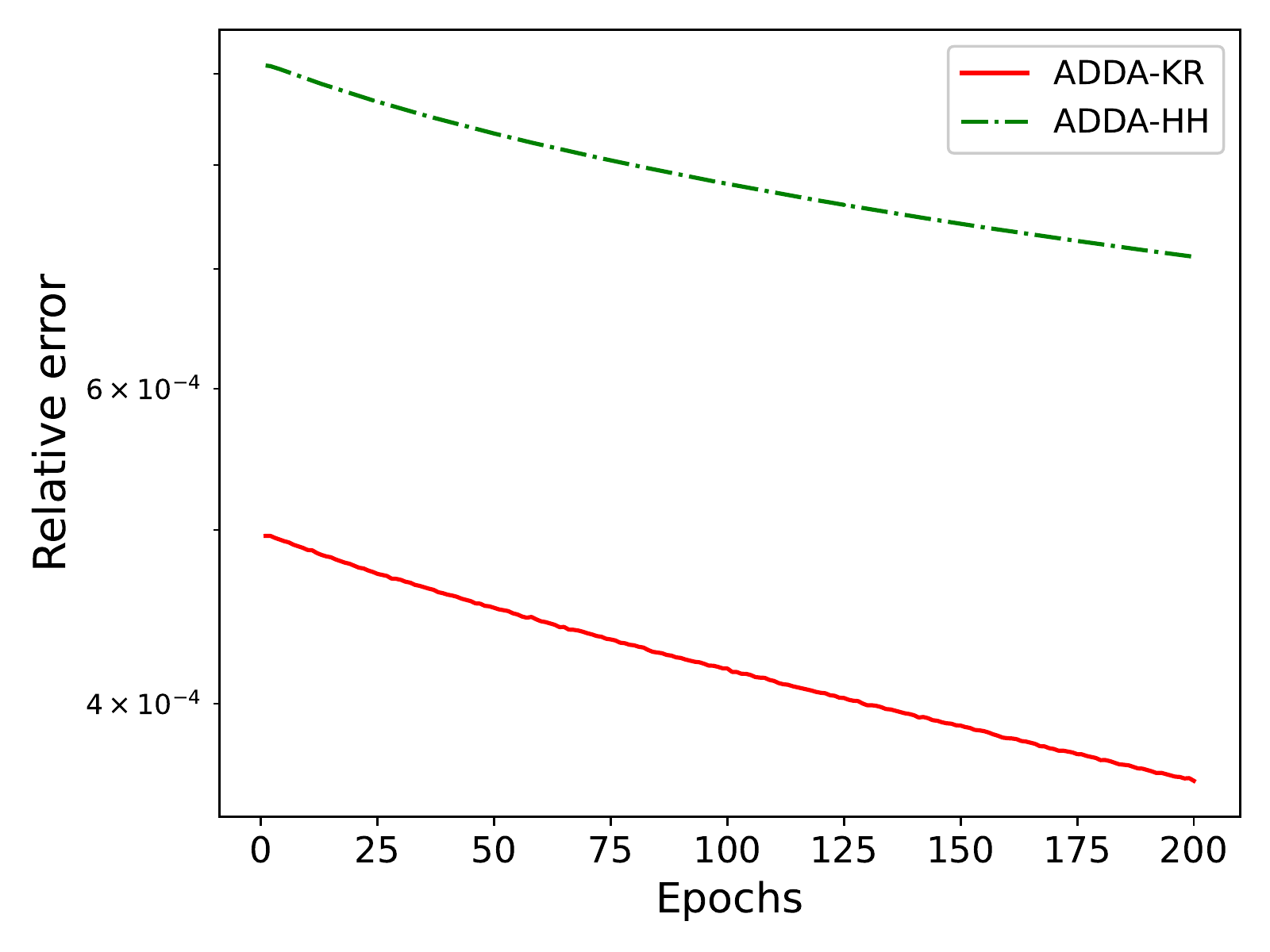}}
	\caption{The relative error for ADDA-KR and ADDA-HH,  two-dimensional bimodal test problem.}
	\label{fig:mix2d}
\end{figure}

\begin{figure}[ht]
     \centering
     \subfloat[][Exact solution $p(\mb{x})$]{\includegraphics[width=.48\textwidth]{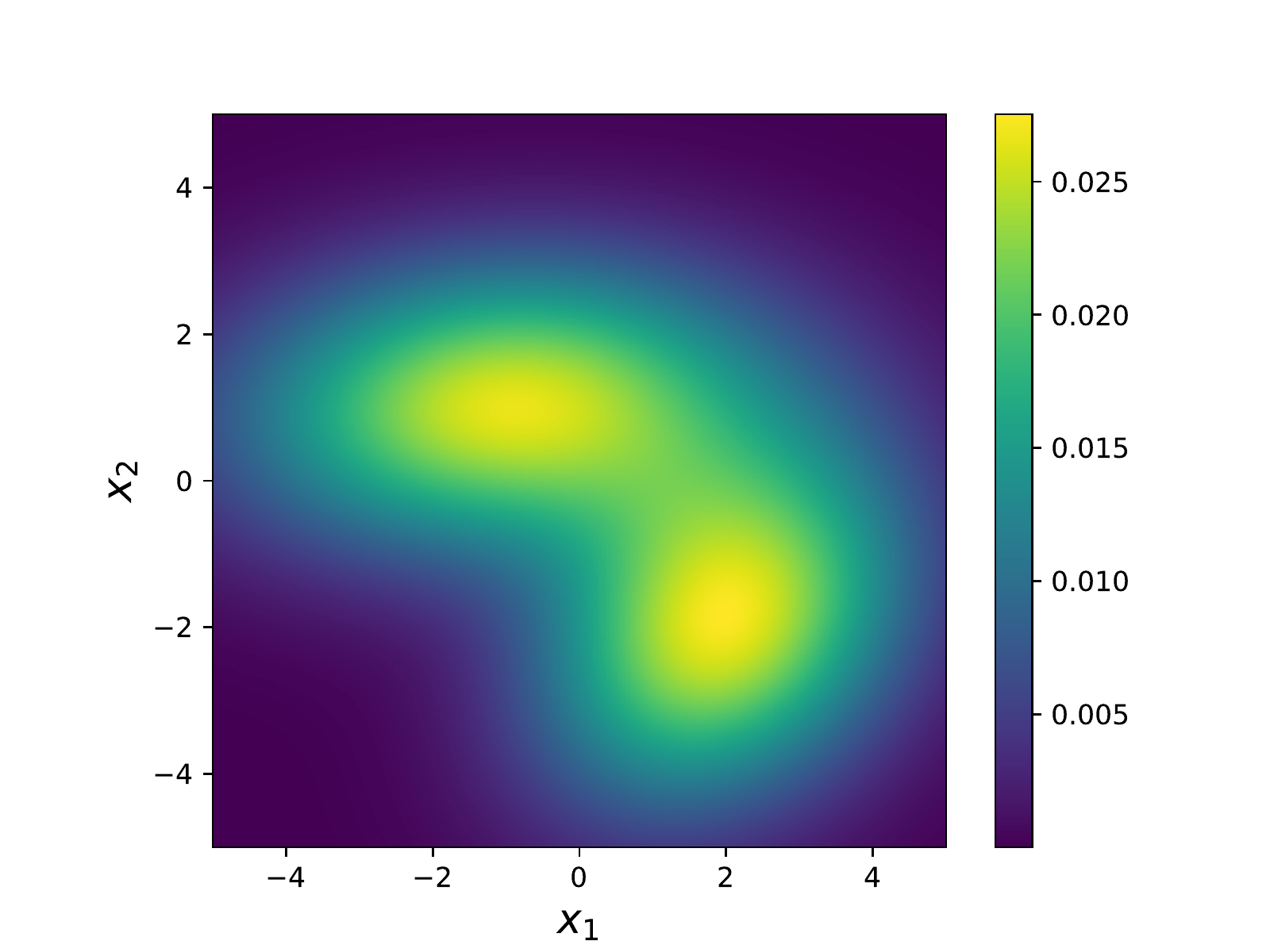}}\quad
     \subfloat[][ADDA-KR solution $p_{X}(\mb{x};\Theta)$]{\includegraphics[width=.48\textwidth]{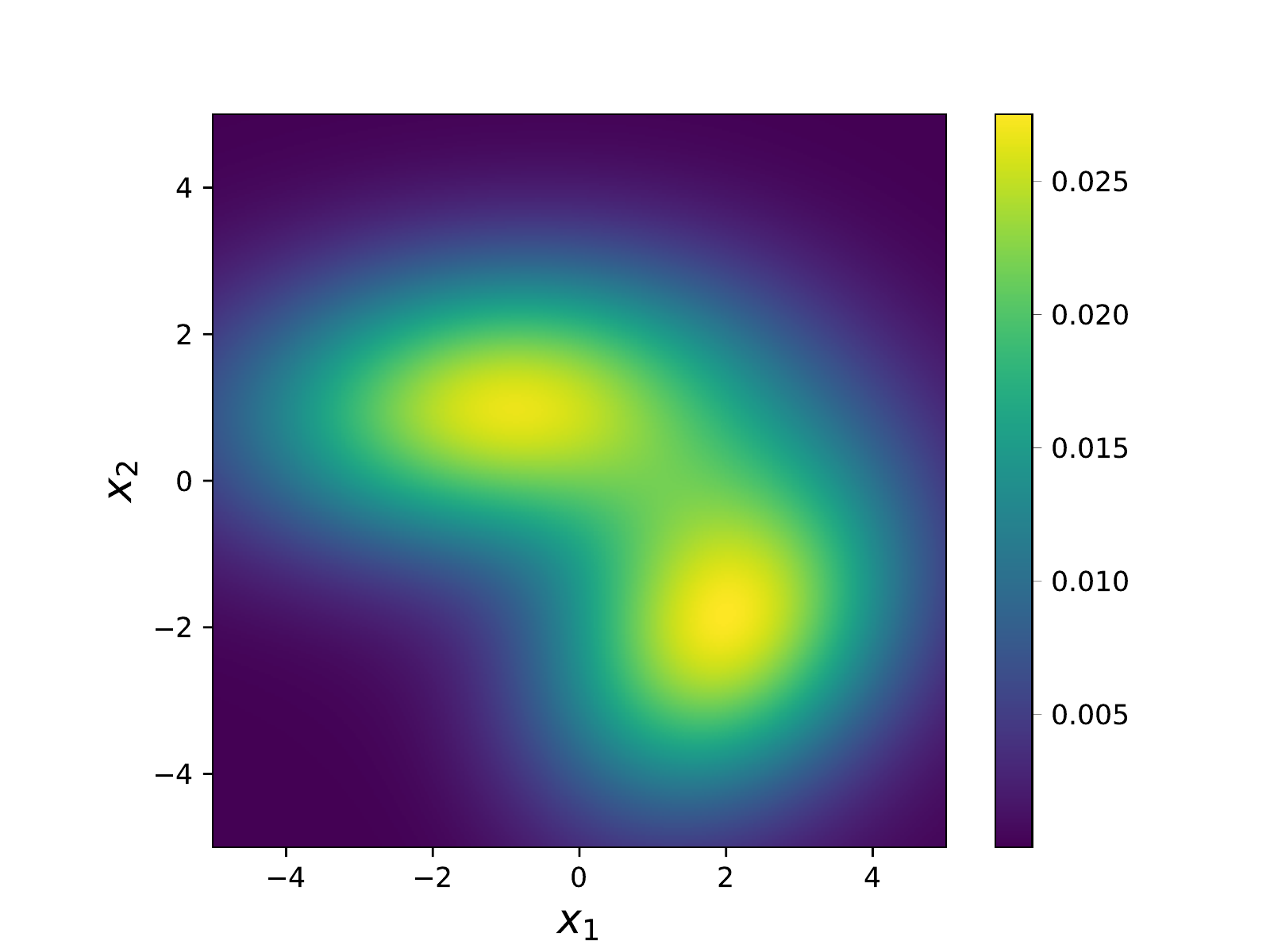}}
     \caption{Solutions, two-dimensional bimodal test problem.}
     \label{fig:mix2d_pdf}
\end{figure}

Figure \ref{fig:mix2d_k_convergence} shows 
the relative error between the exact solution $p(\mb{x})$ and
our ADDA-KR solution $p_X(\mb{x};\Theta)$ at each adaptivity iteration step $k$. 
It is clear that, as the adaptivity iteration step increases, 
the relative error decreases quickly. 
In addition, it can be seen that as the number of epochs increases, the relative error decreases.
Figure \ref{fig:mix2d} shows the comparison between our ADDA-KR and ADDA-HH.
From Figure \ref{fig:mix2d}(a), it can be seen that the relative error of ADDA-KR is smaller than that
of ADDA-HH at each adaptivity iteration step.
Figure \ref{fig:mix2d}(b), Figure \ref{fig:mix2d}(c) and Figure \ref{fig:mix2d}(d) show the relative error 
decreases as the number of epochs increases, at adaptivity iteration steps $k=1,3,5$ respectively.
It can be seen that the relative error of ADDA-KR is clearly smaller than
that of ADDA-HH for each value of epochs, except for the situations that the epoch number is smaller than $125$ at the first adaptivity iteration in \ref{fig:mix2d}(b).  
Figure \ref{fig:mix2d_pdf} shows the exact solution $p(\mb{x})$ and the ADDA-KR solution $p_X(\mb{x};\Theta)$, 
where it can be seen that this bimodal distribution is well approximated by our ADDA-KR solution.


\subsection{High-dimensional bimodal distributions (four-dimensional and eight-dimensional test problems)} \label{sec_test_hdfp}
In this part, we again consider the Fokker-Planck equation with two peaks Eq.\eqref{eq_testfp_bimodal}
and set $\beta_1 = 0.55, \beta_2 = 0.45$.
However, the dimensionality of the problem considered in this part is different from section \ref{sec_test_2dfp}.
We here consider a four-dimensional ($d=4$) problem and an eight-dimensional ($d=8$) problem. 
The exact solution of Eq.\eqref{eq_testfp_bimodal} is a Gaussian mixture distribution Eq.\eqref{eq_bisol}. 
For $d = 4$, we set
\begin{equation}
\begin{aligned}
\mb{\mu}_1 = [-1,-1,-0.3,-0.3]^\mathsf{T}, &\  \mb{\mu}_2 = [2, 2, 0.6,0.6]^\mathsf{T} \\
\mb{\Sigma}_1^{\prime} = \left[ \begin{array}{cc}
         \mb{\Sigma}_1 & \mb{0}  \\
         \mb{0} & 0.6 \mb{\Sigma}_1
\end{array}
\right],  &\quad \mb{\Sigma}_2^{\prime} = \left[ \begin{array}{cc}
         \mb{\Sigma}_2 & \mb{0}  \\
         \mb{0} & 0.6 \mb{\Sigma}_2
\end{array}
\right],
\end{aligned}
\end{equation} 
where $\mb{\Sigma}_1$ and $\mb{\Sigma}_2$ are given in Eq.\eqref{eq_2dmix_setting}, 
and $\mb{\Sigma}_1^{\prime}$ and $\mb{\Sigma}_2^{\prime}$ are the covariance matrices of $p_1$ and $p_2$ for this test problem.

Similarly to the previous settings, we generate  the initial parameters $\Theta_{0}^{(0)}$ with Glorot Gaussian initialization,
and  then construct the initial KRnet $p_X^{(0)}(\mb{x};\Theta_{0}^{(0)})$. 
The number of epochs is set to $N_e=1$, 
and the number of adaptivity iterations conducted for this problem is set to $N_{\rm adaptive}=16$. 
Here, KRnet is trained and sampled in an interleaved manner. 
That is for both the four-dimensional and the eight-dimensional test problems, 
samples $\mathcal{C}_k$ drawn at the $k$-th adaptivity iteration are
immediately used for training KRnet at the $(k+1)$-th iteration, 
while $p_{X}^{(k+1)}(\mb{x};\Theta)$ is immediately used for 
sampling.
The learning rate for Adam optimizer is set to $\eta=0.0001$, 
and the batch size is set to $m = 500$. 
The initial training set $\mathcal{C}_0$ is generated through the uniform distribution with range $[-6,6]^4$, 
and two cases of the collocation sample size are considered: one is $10^5$ and the other is $2\times10^5$.
In addition, we take $L = 8$ affine coupling layers for KRnet, and $L = 16$ for real NVP. The architecture of $\mathsf{NN}_{[i]}$ is the same as that shown in Figure \ref{fig:nn_affine_coupling} with $w = 120$. For KRnet, we set $K=3$. The rotation layer and the nonlinear layer are turned on. To assess the accuracy of ADDA-KR, we again compute the relative error Eq.\eqref{eq_relative_error} between $p(\mb{x})$ and $p_X(\mb{x};\Theta)$ using $3.2 \times 10^5$ validation samples drawn from the exact solution.

\begin{figure}[!ht]
	\centering
	\subfloat[][ADDA-KR and ADDA-HH]{\includegraphics[width=.2785\textwidth,height=.2470\textwidth]{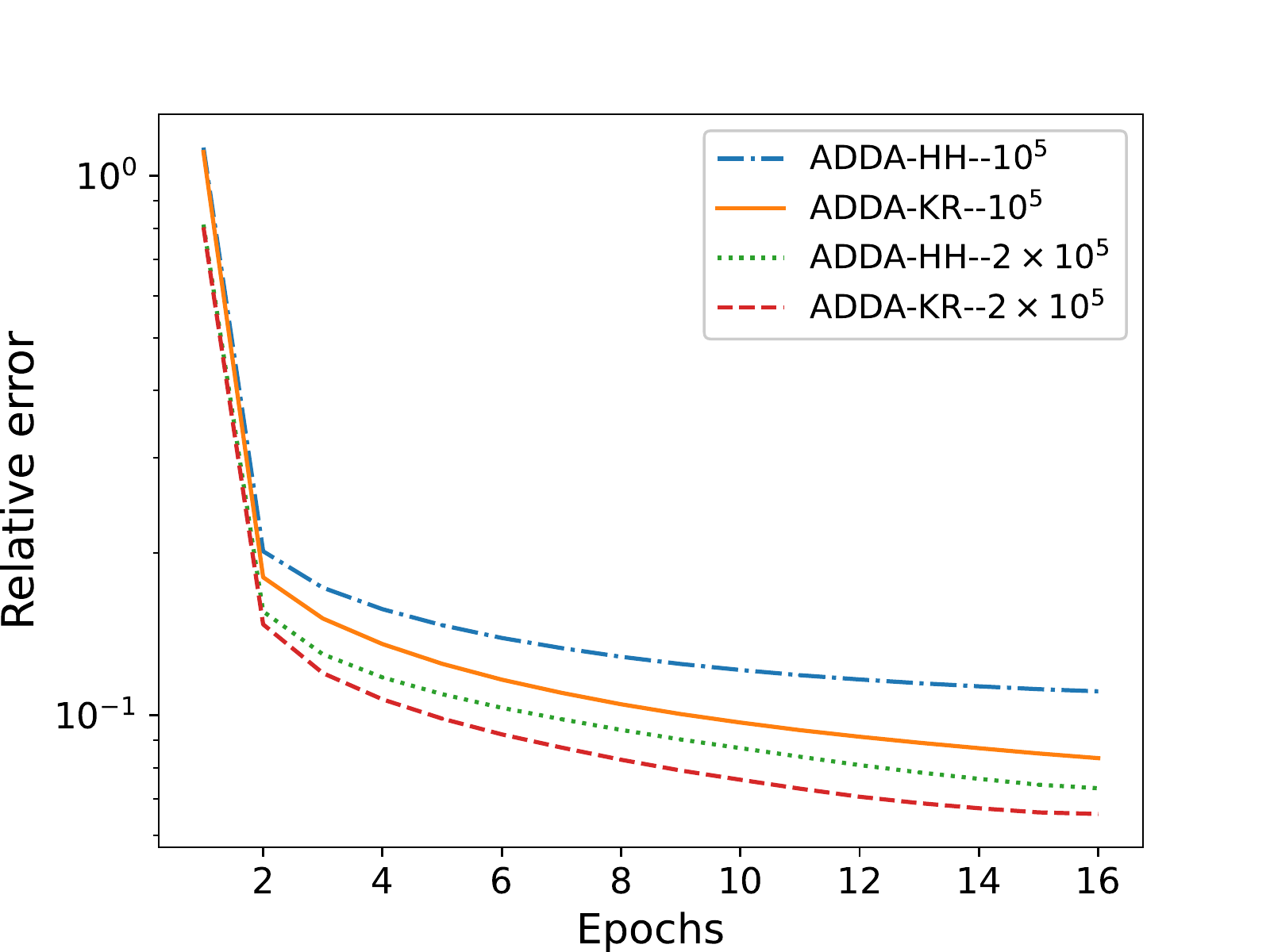}}\quad
	\subfloat[][ADDA-HH  and Uniform-HH]{\includegraphics[width=.2785\textwidth,height=.2470\textwidth]{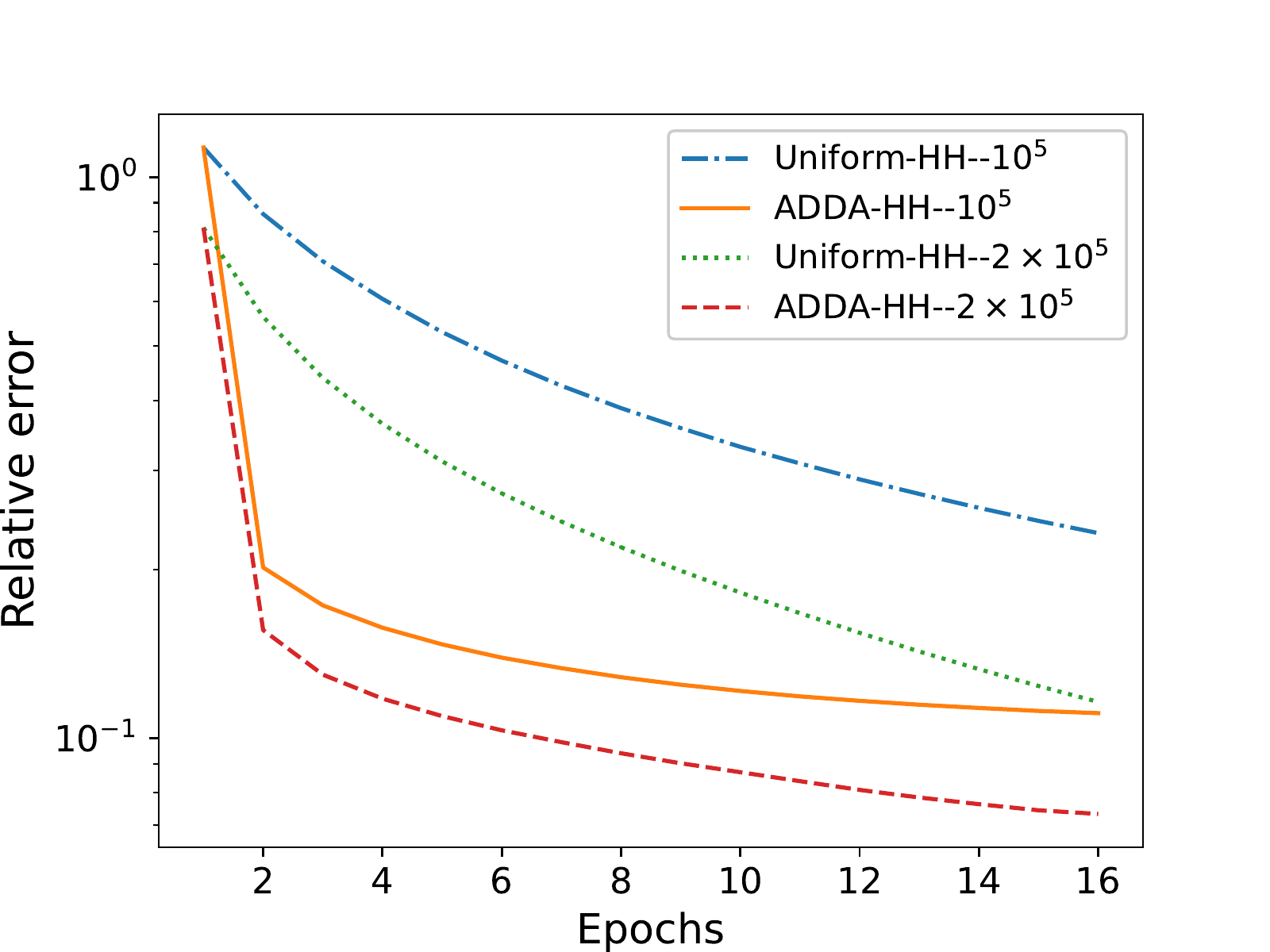}}\quad
	\subfloat[][ADDA-KR  and Uniform-KR]{\includegraphics[width=.2785\textwidth,height=.2470\textwidth]{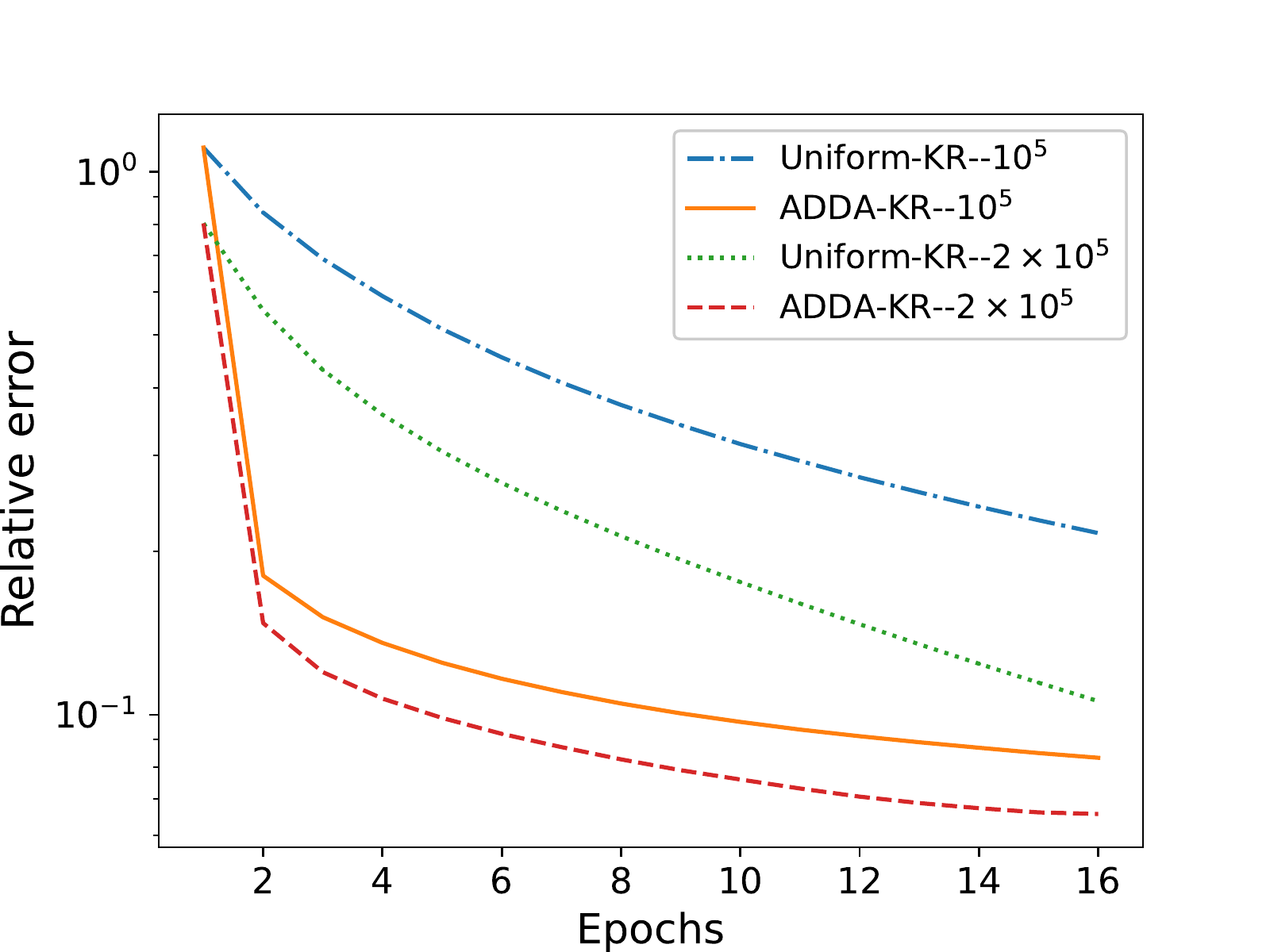}}\\
	\caption{Relative errors, four-dimensional test problem.}
	\label{fig_kl_4d}
\end{figure}

Figure \ref{fig_kl_4d} shows the relative error between $p(\mb{x})$ and $p_X(\mb{x};\Theta)$ for ADDA-KR and ADDA-HH, 
where different numbers of collocation points are considered.  
From Figure \ref{fig_kl_4d}(a), it can be seen that the relative error of ADDA-KR is smaller than that of ADDA-HH. 
From Figure \ref{fig_kl_4d}(b) and Figure \ref{fig_kl_4d}(c), 
as the number of epochs increases, 
the relative errors of ADDA-KR and ADDA-HH decrease quickly, while the relative errors of the uniform sampling strategies (Uniform-KR and
Uniform-HH) decrease slowly. 
In addition, it can be seen that the relative error decreases as the number of training points increases from $10^5$ to $2 \times 10^5$ for ADDA-KR, ADDA-HH, Uniform-KR and Uniform-HH.

Finally, we consider an eight-dimensional bimodal distribution. For this problem, we set
\begin{equation}
\begin{aligned}
\mb{\mu}_1 = [-1,-1,-0.3,-0.3, -0.4, -0.4, -1.6, -1.6]^\mathsf{T}, &\  \mb{\mu}_2 = [2, 2, 0.6,0.6, 0.8, 0.8,2.3,2.3]^\mathsf{T} \\
\tilde{\mb{\Sigma}}_1 = \left[ \begin{array}{cccc}
         \mb{\Sigma}_1 & \mb{0} & \mb{0} & \mb{0}  \\
         \mb{0} & 0.6 \mb{\Sigma}_1 & \mb{0} & \mb{0} \\
         \mb{0} & \mb{0} & 0.8 \mb{\Sigma}_1 & \mb{0} \\ 
         \mb{0} & \mb{0} & \mb{0} & 1.2 \mb{\Sigma}_1
\end{array}
\right],  &\quad \tilde{\mb{\Sigma}}_2 = \left[ \begin{array}{cccc}
         \mb{\Sigma}_2 & \mb{0} & \mb{0} & \mb{0}  \\
         \mb{0} & 0.6 \mb{\Sigma}_2 & \mb{0} & \mb{0} \\
         \mb{0} & \mb{0} & 0.8 \mb{\Sigma}_2 & \mb{0} \\ 
         \mb{0} & \mb{0} & \mb{0} & 1.2 \mb{\Sigma}_2
\end{array}
\right],
\end{aligned}
\end{equation} 
where $\mb{\Sigma}_1$ and $\mb{\Sigma}_2$ are given in Eq.\eqref{eq_2dmix_setting},
and $\tilde{\mb{\Sigma}}_1$ and $\tilde{\mb{\Sigma}}_2$ are the covariance matrices of $p_1$ and $p_2$ for this test problem. 

Again, we generate  the initial parameters $\Theta_{0}^{(0)}$ with Glorot Gaussian initialization,
and  then construct the initial KRnet $p_X^{(0)}(\mb{x};\Theta_{0}^{(0)})$. 
The number of epochs is set to $N_e=1$, 
and the maximum number of adaptivity iterations conducted for this problem is set 
to $N_{\rm adaptive}=120$. 
The learning rate for Adam optimizer is set to $\eta=0.0001$, and the batch size is set to $m = 4000$. 
The initial training set $\mathcal{C}_0$ is generated through the uniform distribution with range $[-6,6]^8$,
and two cases of the collocation sample size are considered: one is $3.2\times10^5$ and the other is $6.4\times10^5$.
In addition, we take $L = 10$ affine coupling layers for KR, and $L = 20$ for real NVP. The architecture of $\mathsf{NN}_{[i]}$ is the same as
that  shown in Figure \ref{fig:nn_affine_coupling} with $w = 160$. For KRnet, we set $K=3$. The rotation layer and the nonlinear layer are turned on. 
We again compute the relative error Eq.\eqref{eq_relative_error}   
using $3.2 \times 10^5$ validation samples drawn from the exact solution.

\begin{figure}[!ht]
	\centering
	\subfloat[][ADDA-KR and ADDA-HH]{\includegraphics[width=.2785\textwidth,height=.2470\textwidth]{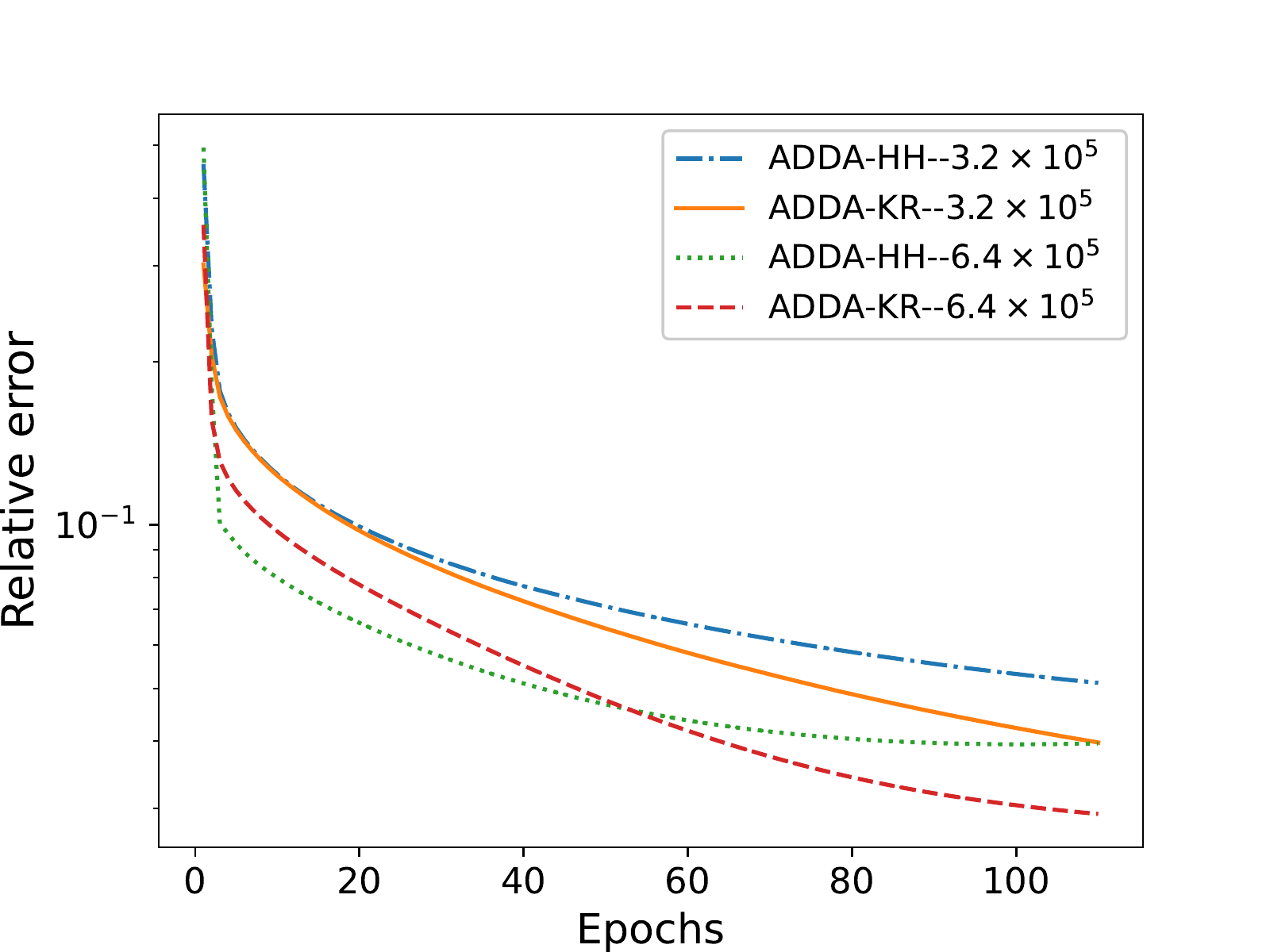}}\quad
	\subfloat[][ADDA-HH  and Uniform-HH]{\includegraphics[width=.2785\textwidth,height=.2470\textwidth]{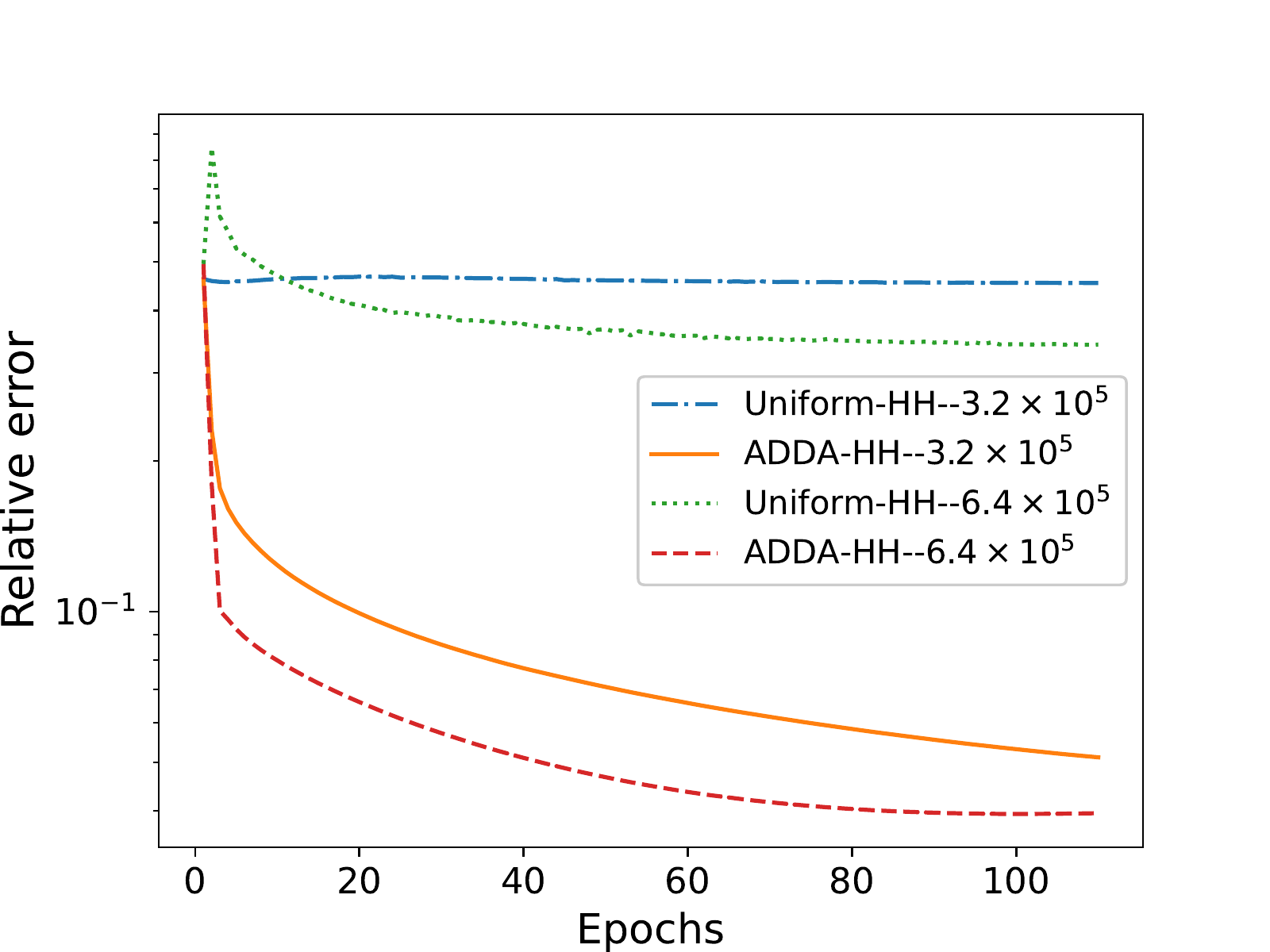}}\quad
	\subfloat[][ADDA-KR  and Uniform-KR]{\includegraphics[width=.2785\textwidth,height=.2470\textwidth]{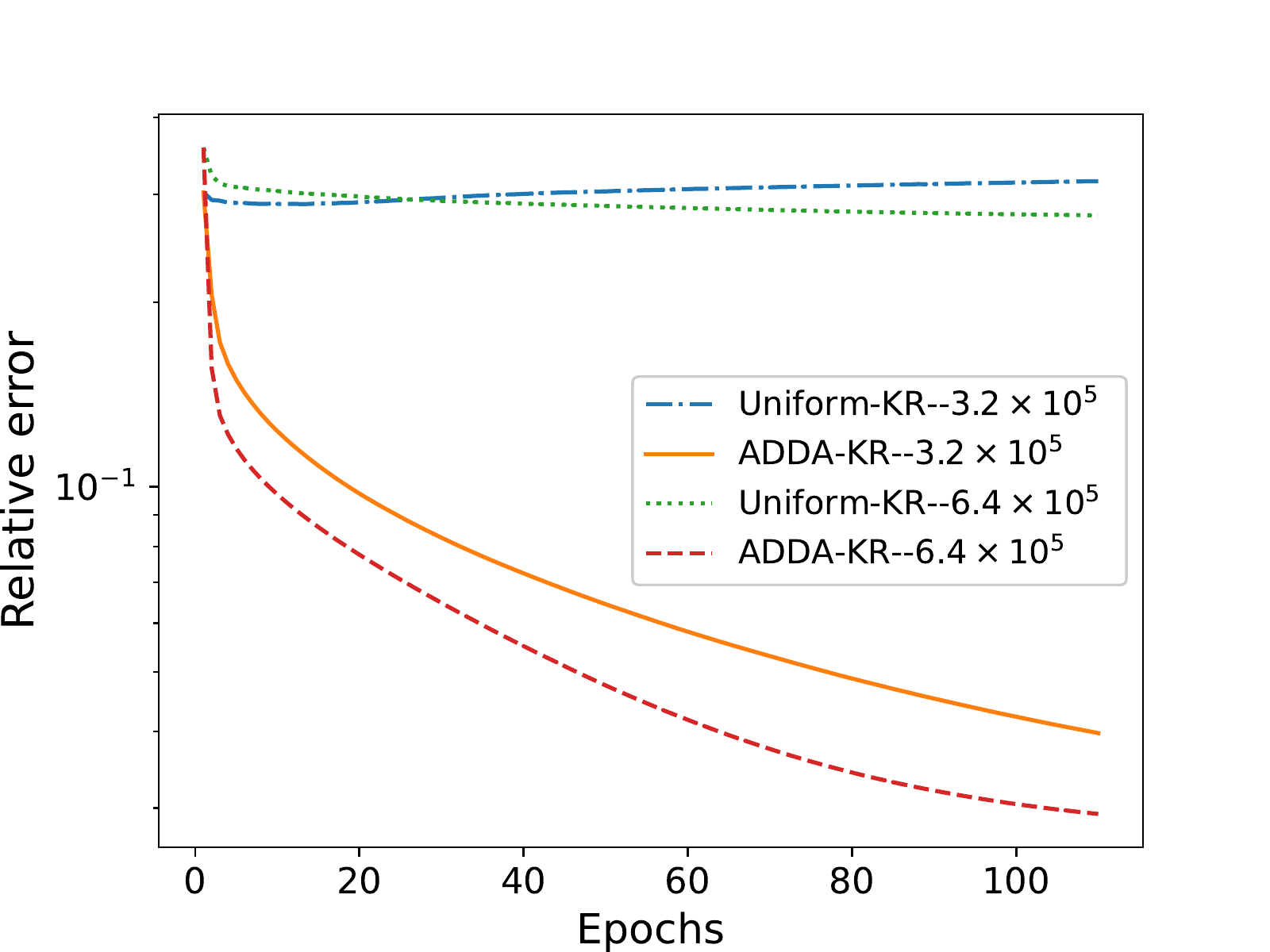}}\\
	\caption{Relative errors, eight-dimensional test problem.}
	\label{fig_kl_8d}
\end{figure}

Figure \ref{fig_kl_8d} shows the relative error between $p(\mb{x})$ and $p_X(\mb{x};\Theta)$ for ADDA-KR and ADDA-HH. 
From Figure \ref{fig_kl_8d}(a), it can be seen that the relative error of ADDA-KR is smaller than that of ADDA-HH, when the
number of epochs is larger than $60$.  
From Figure \ref{fig_kl_8d}(b) and Figure \ref{fig_kl_8d}(c), 
as the number of epochs increases, 
the relative errors of ADDA-KR and ADDA-HH decrease quickly, while the relative errors of the uniform sampling strategies (Uniform-KR and
Uniform-HH) decrease slowly. 
In addition, it can be seen that the relative error decreases as the number of collocation points increases from $3.2 \times 10^5$ to $6.4 \times 10^5$ for ADDA-KR, ADDA-HH, Uniform-KR and Uniform-HH.


